%% file: TIP.tex
\input{Def}

\documentclass[journal]{IEEEtran}
\usepackage{amsmath,amssymb} 
\usepackage{tikz}
\usepackage{graphicx}
\usepackage{comment}
\usepackage{color}

\usepackage{hyperref}
\usepackage{epsfig}
\usepackage[ruled,norelsize]{algorithm2e}
\usepackage[normalem]{ulem}
\usepackage{subfigure}
\usepackage{mathtools}
\usepackage{wrapfig}

\usepackage{times}
\usepackage{epsfig}
\usepackage{graphicx}
\usepackage{amsmath}
\usepackage{amssymb}

\usepackage{rotating}
\usepackage{multirow}
\usepackage{setspace}
\usepackage{algorithm2e}
\usepackage{fancyhdr} 
 \usepackage{color}
\usepackage{array}
\usepackage{subfigure}
\usepackage{soul}
\usepackage{verbatim} 
\usepackage{paralist}
\usepackage{xspace}
\usepackage{booktabs}
\usepackage{enumitem}
\usepackage{kotex}
\usepackage{microtype}

\DeclareMathOperator*{\argmax}{argmax}

\ifCLASSINFOpdf
\else
\fi
\begin{document}
%
\title{Semi-Supervised Image Captioning by Adversarially Propagating Labeled Data}
%
%
%

\author{Dong-Jin Kim,~\IEEEmembership{Member,~IEEE,}
        Tae-Hyun Oh,~\IEEEmembership{Member,~IEEE,}\\
        Jinsoo Choi,~\IEEEmembership{Member,~IEEE,}
        In So Kweon,~\IEEEmembership{Member,~IEEE.}
\IEEEcompsocitemizethanks{\IEEEcompsocthanksitem D.-J. Kim, is with the Department of Data Science, Hanyang University, Seoul, Republic of Korea.\protect\\
E-mail: djdkim@hanyang.ac.kr
\IEEEcompsocthanksitem T.-H. Oh is with the Department of Electrical Engineering and Graduate School of AI (GSAI), POSTECH, Pohang, Republic of Korea (Joint affiliated with Yonsei University).\protect\\ 
E-mail: \url{taehyun@postech.ac.kr}
\IEEEcompsocthanksitem J. Choi and I. S. Kweon are with the School of Electrical Engineering, KAIST, Daejeon, Republic of Korea.\protect\\
E-mail: \{jinsc37,iskweon77\}@kaist.ac.kr
}
\thanks{}}

%
%

\markboth{Journal of \LaTeX\ Class Files,~Vol.~14, No.~8, August~2015}%
{Shell \MakeLowercase{\textit{et al.}}: Bare Demo of IEEEtran.cls for IEEE Journals}
%

\maketitle


\begin{abstract}
We present a novel data-efficient \emph{semi-supervised} framework to improve the generalization of image captioning models.
Constructing a
large-scale labeled image captioning dataset is an expensive task in terms of labor, time, and cost.
In contrast to manually annotating all the training samples,
separately
collecting uni-modal datasets is immensely easier, \eg~a large-scale image dataset and a sentence dataset.
We leverage such massive \emph{unpaired} image and caption data upon standard
paired data by learning to associate them. 
To this end, our proposed semi-supervised learning method assigns pseudo-labels to unpaired samples in an adversarial learning fashion, where the joint distribution of image and caption is learned.
Our method trains a captioner to learn from a paired data and to progressively associate unpaired data.
This approach
shows noticeable performance improvement
even in challenging scenarios including out-of-task data (\ie~relational captioning, where the target task is different from the unpaired data) and web-crawled data.
We also show that our proposed method is theoretically well-motivated and has a favorable global optimal property.
Our extensive and comprehensive empirical results 
both on 
(1) image-based and (2) dense region-based captioning datasets 
followed by comprehensive analysis on the
scarcely-paired COCO dataset
demonstrate the consistent effectiveness of our semi-supervised learning method with unpaired data compared to competing methods.
\end{abstract}


\IEEEpeerreviewmaketitle

\section{Introduction}\label{sec:introduction}
\input{1_sec_intro_TMM}

\section{Related Work}

\input{2_sec_related_TMM}
\section{Proposed Method}
\input{3_sec_method_TMM}

\section{Experiments}

\input{4_sec_experiment_TMM}


\section{Conclusion}
We introduce a method to train an image captioning model with a large scale unpaired image and caption data
upon typical paired data. 
Our framework achieves favorable performance compared to various methods and setups. 
Unpaired captions and images are the data that can be easily collected from the web.
It can also facilitate application-specific captioning models, where labeled data is scarce. 

Before concluding our work, we discuss potential directions to further improve our method.
While the theoretical analysis does not directly provide how to improve, 
an interesting implication of the analysis is that, while our problem setting and modeling are notably different from GAN, the result is seamlessly boiled down to a similar form of that of GAN.
This suggests that one may exploit this analogy between our method and GAN to further improve our method.
This would be crucial ingredients to stimulate creative follow-up research.



\ifCLASSOPTIONcaptionsoff
  \newpage
\fi

{\small
\bibliographystyle{ieee}
\bibliography{egbib}
}

\end{document}


%
\title{Semi-Supervised Image Captioning by Adversarial Learning to Associate Unpaired Data}
%
%
%

\author{Dong-Jin Kim,~\IEEEmembership{Member,~IEEE,}
        Tae-Hyun Oh,~\IEEEmembership{Member,~IEEE,}\\
        Jinsoo Choi,~\IEEEmembership{Member,~IEEE,}
        In So Kweon,~\IEEEmembership{Member,~IEEE.}
\IEEEcompsocitemizethanks{\IEEEcompsocthanksitem D.-J. Kim, is with the Department of Data Science, Hanyang University, Seoul, Republic of Korea.\protect\\
E-mail: djdkim@hanyang.ac.kr
\IEEEcompsocthanksitem T.-H. Oh is with the Department of Electrical Engineering and Graduate School of AI (GSAI), POSTECH, Pohang, Republic of Korea (Joint affiliated with Yonsei University).\protect\\ 
E-mail: \url{taehyun@postech.ac.kr}
\IEEEcompsocthanksitem J. Choi and I. S. Kweon are with the School of Electrical Engineering, KAIST, Daejeon, Republic of Korea.\protect\\
E-mail: \{jinsc37,iskweon77\}@kaist.ac.kr
}
\thanks{}}

%
%

\markboth{Journal of \LaTeX\ Class Files,~Vol.~14, No.~8, August~2015}%
{Shell \MakeLowercase{\textit{et al.}}: Bare Demo of IEEEtran.cls for IEEE Journals}
%

\renewcommand{\theequation}{A.\arabic{equation}}

\newcommand{\AEref}[1]{AEq.~(\ref{#1})}

 \clearpage
\section*{Appendix -- Proofs}
\input{5_sec_appendix}

\IEEEpeerreviewmaketitle



\ifCLASSOPTIONcaptionsoff
  \newpage
\fi

{\small
\bibliographystyle{ieee}
\bibliography{egbib}
}

%% file: 1_sec_intro_TMM.tex

Image captioning is a task of automatically generating a natural language description of a given image.
It is a highly useful for image understanding, in that 1) it extracts the essence
of an image into a self-descriptive form,
and 2) the output format is an interpretable natural language, which is free-form and easy to manipulate so that it can be beneficial to user interactable applications such as language based image 
retrieval~\cite{karpathy2015deep}, video summarization~\cite{choi2018contextually}, navigation~\cite{wang2018look}, and vehicle control~\cite{kim2018textual}.
Image captioning is also general, in that it is not confined to a few number of pre-defined classes.
This enables descriptive analysis of an image.

Recent research on image captioning has made impressive progress~\cite{anderson2017bottom,cornia2020meshed,vinyals2015show}.
Despite this progress, the majority of works are trained only via supervised learning 
where it would be hard to transfer a model to a target domain with significant domain shift~\cite{chen2017show}.
One way to improve the image captioning model's generalizability would be to add more supervised data, which is hard in practice.
Specifically, the MS COCO caption dataset is constructed with 120,000 number of images that were asked annotators to provide five plausible sentences for each image, which is an expensive task in terms of labor, time, and cost.
Moreover, if the target task is a higher-level task involving multiple captions and bounding boxes per image, it becomes even more challenging to annotate the dataset.
For example, for the relational captioning task~\cite{kim2019dense}, dense and combinatorially associated caption and a pair of two bounding boxes are used as a label, and the data for this task has much higher complexity than that of the standard image captioning task.
Constructing such human-labeled datasets is an immensely laborious and time-consuming task, so that building new datasets according to different needs of target themes or application scenarios would be impractical.
Therefore, our goal is to effectively improve image captioning in a more data efficient way.


\begin{figure}[t]
\centering
   \includegraphics[width=1.0\linewidth]{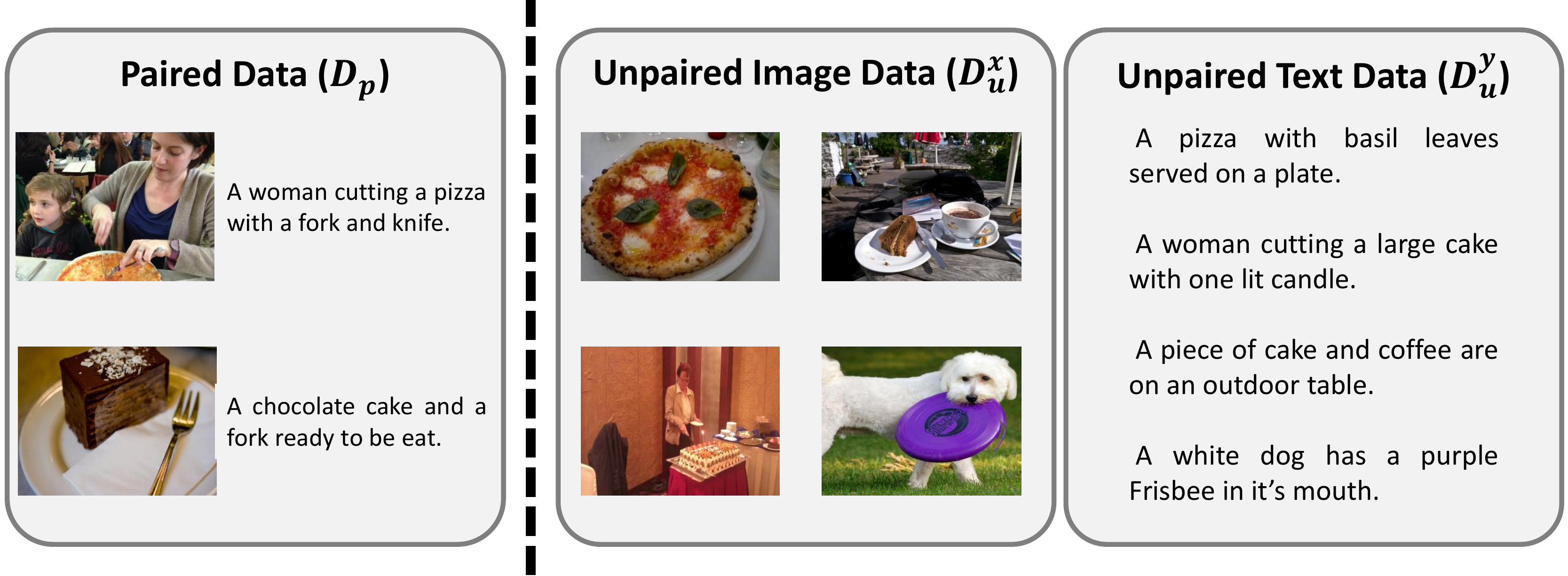}
   \vspace{-5mm}
   \caption{The proposed data setup utilizes 
   ``unpaired'' image-caption data
   upon
   ``paired'' data. We denote  paired data as $\mathcal{D}_{p}$ and unpaired image and caption datasets as $\mathcal{D}_{u}^x$ and $\mathcal{D}_{u}^y$ respectively.
   \vspace{-2mm}
   }
\end{figure}

In this work, we present a novel way of leveraging \emph{unpaired} image and caption data from the web upon traditional elaborately labeled paired data to 
effectively improve
image captioning neural networks.
We are motivated by the fact that images can be easily obtained from the web, and captions can be easily augmented and synthesized by replacing or adding different words for given sentences according to parts of speech as done in~\cite{zhang2015character}.
Moreover, given a sufficient amount of descriptive captions, it is easy to crawl \emph{corresponding but noisy} images through Google or Flickr image databases~\cite{thomee2016yfcc100m} to build a large image corpus.
In this way, we can easily construct a large-scale \emph{unpaired} dataset of images and captions, which requires no or minimal human effort.

Due to the unpaired nature of images (input) and captions (output supervision) in our problem, the conventional supervised learning approaches can no longer be directly used.
We propose to algorithmically assign supervision labels, termed as \emph{pseudo-labels}, to make unpaired data paired. 
The pseudo-label is used as a \emph{learned} supervision label.
To develop the mechanism of pseudo-labeling, we are motivated and leverage the generative capability of generative adversarial networks (GAN)~\cite{goodfellow2014generative}, for searching pseudo-labels from unpaired data.
That is, in order to find the appropriate pseudo-labels for unpaired samples, we utilize an adversarial training method for training a discriminator model.
Thereby, the discriminator learns to distinguish between real and fake image-caption \emph{pairs} and to retrieve pseudo-labels as well as to enrich the captioner training.

This work is the extension of Kim~\etal\cite{kim2019image}. 
In this work, we further improve our method with simple yet significantly effective concept transfer technique and analyze our framework by extensively evaluating our method in diverse and challenging scenarios: 
more challenging image captioning baselines~\cite{cornia2020meshed,huang2019attention}, additional caption domain of MS COCO and Flickr~\cite{young2014image}, and the new relational captioning task~\cite{kim2019dense} along with sentence based image retrieval.
Other than empirical results, we also show the theoretical justification of our design of the proposed learning method with respect to a global optimum.

In short, our main contributions are summarized as follows. 
(1) We propose a novel framework for training an image captioner with the unpaired image-caption data 
upon traditional
paired data.
(2) In order to facilitate training with unpaired data, we devise a new semi-supervised learning approach by the novel usage of the GAN discriminator.
In particular, for the scenarios when the number of paired data is scarce, we additionally propose a simple yet effective teacher-student based concept transfer method to leverage an external high-level knowledge to help bridging between unpaired image and caption data in different domains from the paired data.
(3) Beyond the na\"ive image-level captioning task, we extend our method to the relational captioning task in order to demonstrate that our framework can be easily applied to region-based captioning datasets as well with a simple modification.
(4) We link between our practical realization of the proposed learning method and theoretical algorithmic behaviors.
(5) We show the effectiveness of our method through extensive experiments in various challenging setups compared to strong competing methods. 
We demonstrate that, with 60\% of paired data, our method performs comparably with the model supervised by full data, and our method outperforms the competing methods with full paired data.
More surprisingly, our model trained by our learning method with 1\% of paired data plausibly performs well in a qualitative sense.


%% file: 2_sec_related_TMM.tex
{The main goal of our work is to address unpaired image-caption data to improve the generalizability of image captioning.
Therefore, we mainly focus on image captioning and unpaired data handling literature. 
}

\noindent \textbf{Generalizability in Image Captioning.}\quad
Since the introduction of 
the MS COCO dataset~\cite{lin2014microsoft}, image captioning has been extensively studied in computer vision and language community~\cite{anderson2017bottom,cornia2020meshed,hu2022scaling,huang2019attention,kuo2022beyond,nguyen2022grit,pan2020x,rennie2017self,vinyals2015show,xu2015show} 
by virtue of the advancement of deep neural networks~\cite{krizhevsky2012imagenet}.
As neural network architectures become more advanced, \eg~Transformer~\cite{cornia2020meshed,pan2020x}, they require a much larger dataset scale for generalization~\cite{shalev2014understanding} as the image captioning models tend to show limited generalizabiltiy.
Despite the extensive study on network architectures, the data issues such as noisy data, partially missing data, and unpaired data have been relatively less studied in image captioning.

Traditionally, utilizing unpaired image-caption data required additional information
to associate images and captions.
Gu~\etal\cite{gu2018unpaired} introduce additional modal information, Chinese captions, and use it as strong pivot language 
for language pivoting~\cite{utiyama2007comparison}.
Feng~\etal\cite{feng2018unsupervised} propose an unpaired captioning framework which trains a model without either image or sentence labels via learning a visual concept detector with external data, the OpenImage dataset~\cite{krasin2017openimages}.
Laina~\etal\cite{laina2019towards} and Guo~\etal\cite{guo2020recurrent} propose improved training methods given the same visual concept detector as Feng~\etal trained with the OpenImage dataset. We later show that our method can be easily extended with a similar visual concept learning to enhance the performance.
Gu~\etal\cite{gu2019unpaired} utilize scene graph to bridge between unpaired image and caption data.
Chen~\etal\cite{chen2017show} approach image captioning as a domain adaptation by utilizing the large-scale paired MS COCO caption dataset as the source domain and adapting on separate unpaired image or caption datasets
as the target domain.
Kim~\etal\cite{kim2018disjoint} propose multi-task learning method to use an action recognition dataset without caption labels to improve video captioning performance.
Liu~\etal\cite{liu2018show} use self-retrieval rewards for captioning to facilitate training a model with partially labeled data, where the self-retrieval module retrieves corresponding images with the captions generated from the model.
As a separate line of work, there are novel object captioning methods~\cite{anne2016deep,venugopalan2017captioning,lu2018neural} that {additionally} exploit unpaired image and caption data to mine a description of a novel word.


Most of aforementioned works including~\cite{anne2016deep,ben2021unpaired,feng2018unsupervised,gu2018unpaired,gu2019unpaired,guo2020recurrent,kim2018disjoint,laina2019towards,meng2022object} exploit large auxiliary \emph{supervised} datasets such as class labels or scene graph.
To the best of our knowledge, we are the first to study how to handle unpaired image and caption data for image captioning even without any auxiliary information but by leveraging 
semi-supervised
image-caption data only. 
Although Chen~\etal\cite{chen2017show} do not use auxiliary information as well, 
it requires large amounts of paired source data, of which data regime is different from ours. 
Liu~\etal\cite{liu2018show} is also this case, where they use the full paired MS COCO caption dataset with an additional large unlabeled image set.
Our method can deal with those regimes as well as a very scarce paired source data regime, of which scale is only $1\%$ of the COCO dataset at minimum.

\noindent \textbf{Multi-modality in Unpaired Data Handling.}\quad 
By virtue of the advance
on generative modeling,
\eg~GAN~\cite{goodfellow2014generative}, multi-modal translation recently emerged as a popular field.
Among many possible modalities, image-to-image translation between two different and unpaired domains has been  actively explored.
To tackle this problem, the cycle-consistency constraint between unpaired data is exploited in CycleGAN~\cite{zhu2017unpaired} and DiscoGAN~\cite{kim2017learning}, and it is further improved in UNIT~\cite{liu2017unsupervised}.

In this work, we regard image captioning as a multi-modal translation.
Our work has a similar motivation to the unpaired image-to-image translation~\cite{zhu2017unpaired}, unsupervised machine translation~\cite{artetxe2018unsupervised}, and machine translation with monolingual data~\cite{zhang2018joint}.
However, we show that the cycle-consistency does not work on our problem setup due to a significant modality gap.
Instead, our results suggest that the traditional label propagation based semi-supervised framework~\cite{zhou2004learning} is more effective for our task. 



\noindent \textbf{Semi-supervised Learning.}
In general, the goal of semi-supervised learning (SSL) is to improve the model performance by training with unlabeled data under {a transductive assumption}~\cite{chapelle2009semi}. 
Recent deep learning based SSL methods can be divided into four main categories: (1) pseudo-label generation~\cite{lee2013pseudo}, (2) consistency regularization~\cite{miyato2018virtual,tarvainen2017mean}, (3) combination of pseudo-labeling with consistency regularization~\cite{berthelot2019mixmatch,berthelot2020remixmatch,sohn2020fixmatch,kuo2020featmatch}, and (4) generative model based methods~\cite{chongxuan2017triple,gan2017triangle}.
Our method is motivated by the generative model based semi-supervised learning~\cite{chongxuan2017triple}.
While the prior work is mostly limited to deal with simple image classification,
our work extend the regime to 
image and caption modalities.

%% file: 3_sec_method_TMM.tex
\begin{figure*}[t]
\begin{center}
   \includegraphics[width=.8\linewidth]{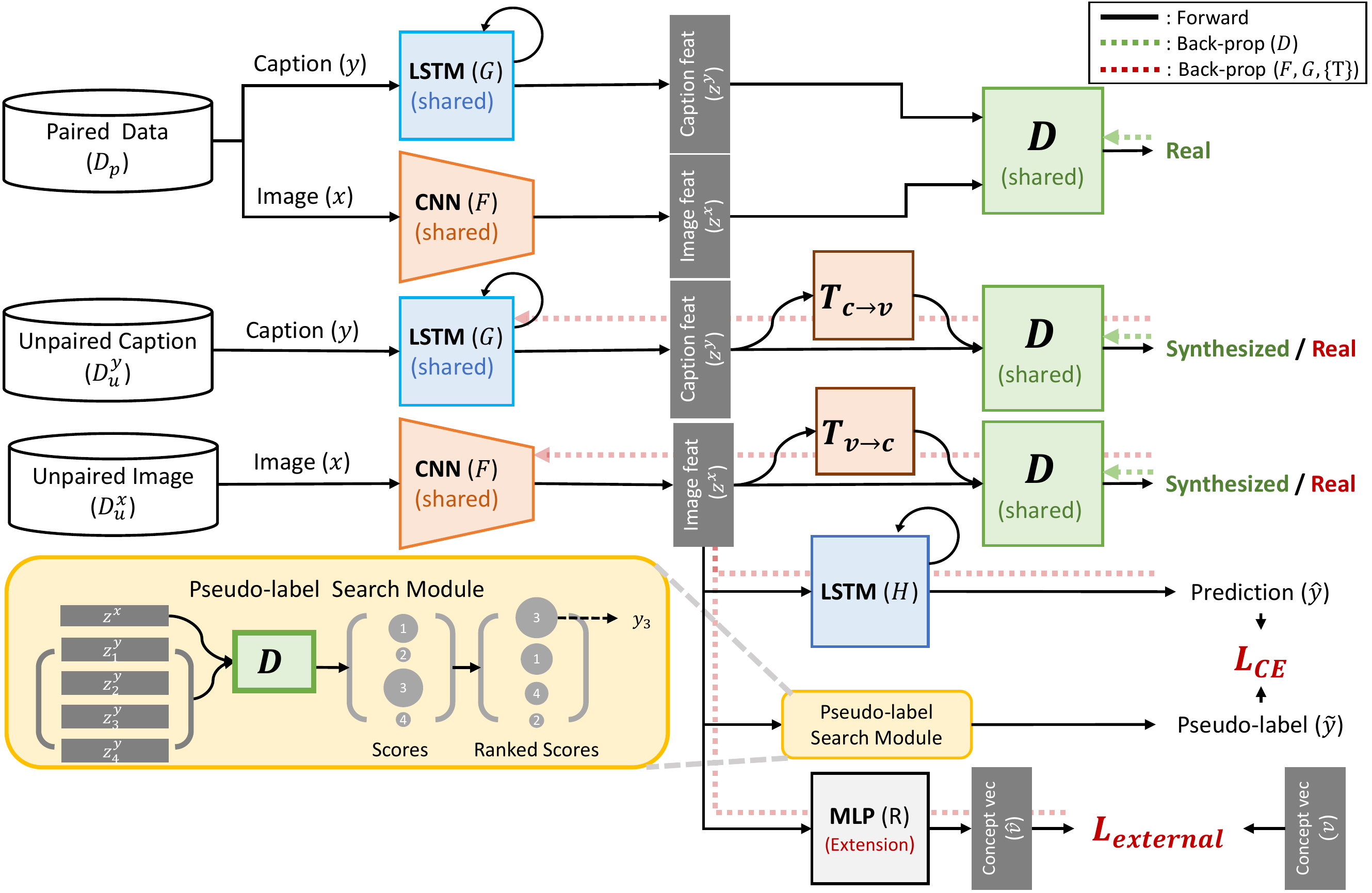}
\end{center}
   \vspace{-2mm}
   \caption{Illustration of the proposed method. Dotted arrows denote the path of the gradients via back-propagation. Given any image and caption pair, CNN and RNN (LSTM) encoders encode input image and caption into the respective feature spaces. A discriminator ($D$) is trained to discriminate whether the given feature pairs are real or fake, while the encoders are trained to fool the discriminator. 
   The learned discriminator is also used to assign the most likely pseudo-labels to unpaired samples through the pseudo-label search module. We additionally introduce an auxiliary multi-layer perceptron to learn external knowledge via concept transfer.
   }
   \vspace{-2mm}
   \label{fig:architecture}
\end{figure*}

In this section, 
we first brief the standard image caption learning and describe how we can leverage the unpaired dataset.
Then, we introduce an adversarial learning method
for obtaining a GAN model that encourages to match the distribution of latent features of images and captions.
The GAN model is used for assigning pseudo-labels, which allows a challenging semi-supervised learning with 
both labeled and unlabeled
data.
Moreover, we analyze the theoretical properties of our proposed framework. 
Lastly, we extend our method to the relational captioning scenario.



\subsection{Adversarial Semi-supervised Training}
Let us denote a dataset with $N_p$ image-caption pairs as $\mathcal{D}_{p} = \{(\mathbf{x}_i, y_i)\}^{N_p}_{i=1}$. 
A typical image captioning task is defined as follows: 
given an image $\textbf{x}_i$, the model generates a caption $y_i$ that best describes the image. 
Traditionally, a captioning model is trained on a large paired dataset $(\mathbf{x},y)\in \mathcal{D}_{p}$, \eg~the MS COCO dataset, 
by minimizing the negative log likelihood against the ground truth caption as follows: 
\begin{equation}
    \centering
    \sum^{}_{(\mathbf{x}, y) \in \mathcal{D}_{p}} L_\textsf{CE}(y,\hat{y}(\mathbf{x})),
\label{eqn:cross-entropy}
\end{equation}
where 
$L_\textsf{CE}$ 
denotes the cross entropy loss,
and $\hat{y}(\mathbf{x})$ denotes
output of the model.
Motivated by 
early neural machine translation literature~\cite{cho2014learning}, 
captioning frameworks have been 
typically implemented as an encoder-decoder architecture~\cite{vinyals2015show}, \ie~CNN-RNN.
The CNN encoder $F(\mathbf{x})$ outputs a latent feature vector $\mathbf{z}^x$ from a given input image $\mathbf{x}$, followed by the RNN decoder $H(\mathbf{z}^x)$ to generate a caption $y$ from $\mathbf{z}^x$ in a natural language form, \ie~$\hat{y}(\mathbf{x}) {=} p(y|\mathbf{x};F,H) {=} H {\circ} F(\mathbf{x})$, as depicted in \Fref{fig:architecture}.


\vspace{1mm}
\noindent\textbf{Learning with Unpaired Data.}
Our problem deals with unpaired data, where the image and caption sets $\mathcal{D}_u^x {=} \{\mathbf{x}_i\}^{N_x}_{i=0}$ and $\mathcal{D}_u^y{=}\{y_i\}^{N_y}_{i=0}$ are not paired.
Given the unpaired datasets, 
due to 
missing annotations,
the loss in \Eref{eqn:cross-entropy} cannot be directly computed.
Motivated by the semi-supervised framework~\cite{shi2018transductive},
we artificially generate \emph{pseudo-labels} for respective unpaired datasets, so that the supervision loss in \Eref{eqn:cross-entropy} can be leveraged with unpaired data.

Specifically, we retrieve the best matched caption $\tilde{y}_i$ in $\mathcal{D}_u^y$ given a query image $\mathbf{x}_i$, 
assign it as a pseudo-label, and vice versa ($\tilde{\mathbf{x}}_i$ for $y_i$).
We express
the pseudo-labeling as a function for simplicity, \ie~$\tilde{y}_i = \tilde{y}(\mathbf{x}_i)$.
To retrieve a semantically meaningful match, we need a measure to assess proper matches.
We use a discriminator network to determine real or fake pairs in a similar way to 
GANs, which will be described in later sections.
With the retrieved pseudo-labels, now we can compute \Eref{eqn:cross-entropy} with unpaired data
as:
\begin{equation}
\centering
\small
    \min_{F,H}
    \lambda_x\sum^{}_{\mathbf{x} \in \mathcal{D}_{u}^x} 
    L_\textsf{CE}(\tilde{y}(\mathbf{x}),\hat{y}(\mathbf{x})) + 
    \lambda_y\sum^{}_{y \in \mathcal{D}_{u}^y} 
    L_\textsf{CE}(y,\hat{y}(\tilde{\mathbf{x}}(y))),
\label{eqn:cross-entropy2}
\end{equation}
where $\lambda_{\{\cdot\}}$ denote the balance parameters.


\vspace{1mm}
\noindent\textbf{Discriminator Learning by Unpaired Feature Matching.}
We train via a
criterion to find a semantically meaningful match, so that pseudo-labels for each modality are effectively retrieved.
To this end, we 
pre-train 
a
discriminator with the paired supervised dataset, and then jointly train it with the other network parts on both paired and unpaired datasets.

We introduce
a caption encoder, $G(y)$, which embeds the caption $y$ into a feature $\mathbf{z}^y$.
This is implemented with a single layer LSTM, and we take the output of the last time step as the caption representation $\mathbf{z}^y$.
Likewise, 
given an image $\mathbf{x}$, we obtain $\mathbf{z}^x$ by
the image encoder $F(\mathbf{x})$.
Now, we have a comparable feature space of $\mathbf{z}^x$ and $\mathbf{z}^y$, of which the number of dimensions are set to be the same.
We train
the discriminator to distinguish whether the pair $(\mathbf{z}^x, \mathbf{z}^y)$ comes from true paired data $(\mathbf{x}, y)\in \mathcal{D}_p$, \ie~the pair belongs to the real distribution $p(\mathbf{x},y)$ or not.

To train the discriminator, we could use random data of $\mathbf{x}$ and $y$ independently sampled from respective unpaired datasets, but we found that
it is detrimental to performance due to uninformative pairs in training.
Instead, we conditionally synthesize $\mathbf{z}^x$ or $\mathbf{z}^y$, {to form a} synthesized pair that appears to be as realistic as possible.
We use the feature transformer networks $\tilde{\mathbf{z}}^y\,{=}\,T_{v\rightarrow c}(\mathbf{z}^x)$ and $\tilde{\mathbf{z}}^x\,{=}\,T_{c\rightarrow v}(\mathbf{z}^y)$, where $v{\rightarrow}c$ denotes the mapping from visual data to caption data and vice versa, and  $\tilde{\mathbf{z}}^{(\cdot)}$ denotes the conditionally synthesized feature.
$\{T\}$ are implemented with multi-layer-perceptron with four fully-connected (FC) layers with the ReLU nonlinearity.

The discriminator $D(\cdot,\cdot)$ learns to distinguish features if they are real or not. 
At the same time, the other associated networks $F$, $G$, $T_{\{\cdot\}}$ are learned to fool the discriminator by matching the distribution of paired and unpaired data.
We formulate 
this adversarial training as follows: 
\begin{equation}
\centering
\small
\begin{aligned}
    &\min_{F,G, \{T\}} \max_{D}  \tilde{U}(F, G, \{T\},D)\\
    =&\min_{F,  G, \{T\}} \max_{D}  U(F, G, \{T\},D) + \mathop{\mathbb{E}}\limits_{\substack{(\mathbf{z}^x, \mathbf{z}^y) \\ \sim (F, G)\circ \mathcal{D}_p}}[ L_{reg}(\mathbf{z}^x, \mathbf{z}^y, \{T\})],
\end{aligned}
\label{eqn:gan}
\end{equation}
where
\begin{equation}
\small
\begin{aligned}
U(F, G, \{T\},D)  =&\mathop{\mathbb{E}}\limits_{\substack{(\mathbf{z}^x, \mathbf{z}^y) \\ \sim (F, G)\circ \mathcal{D}_p}}[\log(D(\mathbf{z}^x,\mathbf{z}^y))] \\
    &+ \tfrac{1}{2}\left( \mathop{\mathbb{E}}\limits_{\mathbf{x} \sim p (\mathbf{x})}[\log(1-D(F(\mathbf{x}),T_{v \rightarrow c}(F(\mathbf{x}))))]\right. \\
    &\left. + \mathop{\mathbb{E}}\limits_{y\sim p (y)}[\log(1-D(T_{c \rightarrow v}(G(y)),G(y)))] \right),
    \end{aligned}
    \label{eq:U}
\end{equation}
{\small{$
L_{reg}(\mathbf{z}^x, \mathbf{z}^y, \{T\}){=}
\lambda_{reg}( \|\mathop{T}\limits_{v \rightarrow
c}(\mathbf{z}^x) {-} \mathbf{z}^y\|_F^2 {+} \|\mathbf{z}^x {-} \mathop{T}\limits_{c \rightarrow v}(\mathbf{z}^y)\|_F^2).
$}}
Note that the first log term in \Eref{eq:U} is not used for updating any learnable parameters related to $F, G,\{T\}$, but only used for updating $D$.
The overall architecture related to this formulation is illustrated in \Fref{fig:architecture}.

Through alternating training of the discriminator ($D$) and generators ($F,G,\{T\}$), the latent feature distribution of paired and unpaired data should be close to each other, \ie~$p(\mathbf{z}^x,\mathbf{z}^y)
{\,\approx\,} p_{v\rightarrow c}(\mathbf{z}^x,\mathbf{z}^y)
{\,\approx\,} p_{c\rightarrow v}(\mathbf{z}^x,\mathbf{z}^y)$, where $p_{v\rightarrow c}(\mathbf{z}^x,\mathbf{z}^y) = p(\mathbf{z}^x)p_{v\rightarrow c}(\mathbf{z}^y|\mathbf{z}^x)$, 
$p_{c\rightarrow v}(\mathbf{z}^x,\mathbf{z}^y) = p(\mathbf{z}^y)p_{c\rightarrow v}(\mathbf{z}^x|\mathbf{z}^y)$, 
and
$p_{v\rightarrow c}(\mathbf{z}^y|\mathbf{z}^x)$ and $p_{c\rightarrow v}(\mathbf{z}^x|\mathbf{z}^y)$ are modeled with $T_{v\rightarrow c}$ and $T_{c\rightarrow v}$, respectively.
It implies that, as the generator is trained, the decision boundary of the discriminator tightens;
hence, we can use the plausibly learned $D$ to retrieve a proper pseudo-label, if the unpaired datasets are sufficiently large such that semantically meaningful matches exist between the different modality datasets.



\vspace{1mm}
\noindent\textbf{Pseudo-label Assignment.}
Given an image $\mathbf{x}\in \mathcal{D}_u^x$, we retrieve a caption in the unpaired dataset, \ie~$\tilde{y} \in \mathcal{D}_u^y$, that has the highest score obtained by the discriminator, \ie~the most likely caption to be paired with the given image as
\begin{equation}
\centering
    \tilde{y}_i = \tilde{y}(\mathbf{x}_i) = \argmax_{y\in \mathcal{D}_u^y}\,\, D\left(F(\mathbf{x}_i), G(y) \right),
    \label{eqn:pseudo_x}
\end{equation}
vice versa for unpaired captions:
\begin{equation}
\centering
    \tilde{\mathbf{x}}_i = \tilde{\mathbf{x}}(y_i) = \argmax_{\mathbf{x}\in \mathcal{D}_u^x}\,\, D\left(F(\mathbf{x}), G(y_i) \right).
    \label{eqn:pseudo_y}
\end{equation}

By this retrieval process over all the unpaired datasets, 
we now have fully paired data; \ie~image-caption pairs $\{(\mathbf{x}_i, y_i)\}$ from the paired data and the pairs with pseudo-labels {$\{(\mathbf{x}_j, \tilde{y}_j)\}$ and $\{(\tilde{\mathbf{x}}_k, y_k)\}$} from the unpaired data. 
However, these pseudo-labels are likely to 
be noisy or biased, 
thus treating them equally with the paired ones 
would not be desirable~\cite{oh2022daso}.
Motivated by learning with noisy labels~\cite{lee2017cleannet,wang2018iterative}, we 
re-weigh the data pairs by defining a confidence score for each of the assigned pseudo-labels.
In order to 
obtain the confidence score,
we propose to use the output score from the discriminator as the confidence score, \ie~$\alpha^x_i {=} \hat{D}(\mathbf{x}_i,\tilde{y}_i)$ and $\alpha^y_i {=} \hat{D}(\tilde{\mathbf{x}}_i,{y}_i)$, where we denote $\hat{D}(\mathbf{x}, y) \,{=}\, D(F(\mathbf{x}), G(y))$, and $\alpha\in [0,1]$ due to the sigmoid function at the final layer.
We utilize the confidence scores to assign weights to the unpaired samples.
The final weighted loss $\min_{F,G} \mathcal{L}_{cap}(F,G)$ is defined as follows:\vspace{2mm}

\noindent
\resizebox{!}{2.4mm}{
$
\min\limits_{F,H} \sum\limits_{(\mathbf{x},y) \in \mathcal{D}_{p}} 
    L_\textsf{CE}(y,\hat{y}(\mathbf{x})) +
    \lambda_x\sum\limits_{\mathbf{x} \in \mathcal{D}_{u}^x} \alpha_{(\tilde{y}(\mathbf{x}),\mathbf{x})}^x 
    L_\textsf{CE}(\tilde{y}(\mathbf{x}),\hat{y}(\mathbf{x}))$
}
\begin{equation}
\small
\resizebox{!}{2.5mm}{
    $+\lambda_y\sum\limits_{y \in \mathcal{D}_{u}^y} \alpha_{(y,\tilde{\mathbf{x}}(y))}^y 
    L_\textsf{CE}(y,\hat{y}(\tilde{\mathbf{x}}(y))).$
}
\label{eqn:pseudo_cross_entropy}    
\end{equation}
To ease training further, we add an additional triplet loss to \Eref{eqn:pseudo_cross_entropy}:
\begin{equation}
\begin{aligned}
    &\mathcal{L}_{triplet}(F,H)=\\
    &\sum_{\substack{(\mathbf{x}_p,y_p)\in \mathcal{D}_{p},\\
    \mathbf{x}_u\in \mathcal{D}_{u}^x,
    y_u\in \mathcal{D}_{u}^y}}
    -\log\frac{p(y_p|\mathbf{x}_p;F,H)}{p(y_p|\mathbf{x}_u;F,H)}-\log\frac{p(y_p|\mathbf{x}_p;F,H)}{p(y_u|\mathbf{x}_p;F,H)},
\end{aligned}
\label{eqn:triplet}
\end{equation}
by regarding random unpaired samples as negative.
This slightly improves the performance.
We jointly train the model on both paired and unpaired data.

\vspace{1mm}
\noindent\textbf{Leveraging External Knowledge via Concept Transfer.}
Although our semi-supervised learning method works properly
to associate unpaired image and caption data despite scarce paired data, the smaller 
the paired data size is,
the more difficult it becomes 
to associate unpaired samples from \emph{unseen} domains.
This is because 
the small paired data lacks the information to capture any snippet of image or text, \ie~concept of each data.
Therefore, as an extension, we propose to borrow a large-scale pre-trained 
knowledge to effectively associate unpaired samples by capturing concept regardless of domain, which is crucial for semi-supervised learning especially when paired data is scarce.



As an external source of knowledge, we propose to use concept embeddings obtained from an off-the-shelf and pre-trained scene understanding model that provides a high-level scene understanding.
We extract a set of dense vectors\footnote{The dense vectors can be any dense representation, \eg~a pixel-wise feature map, feature vectors corresponding to region proposals, \etc.} from an image by using the 
pre-trained model.
By averaging the 
vectors of
the image, we 
obtain a single vector $\mathbf{v} = \text{Concept}(\mathbf{x})$ that represents an image, which we call ``concept vector.''

To borrow knowledge from an external pre-trained model \emph{regardless of its network architecture},
we utilize this concept vector in a way of the knowledge distillation~\cite{hinton2015distilling}, where  
our image encoder $F(\cdot)$ learns the knowledge encoded in the vector.
To make the encoder deal with this auxiliary task, we add an auxiliary concept regression branch $R(\cdot)$ to the penultimate layer.
The auxiliary branch is implemented by a multi-layer perceptron to create a vector $\hat{\mathbf{v}} = R \circ F(\mathbf{x})$ that mimics the concept vector provided by the high-level scene understanding
model.
Then, the image captioning model is trained by adding the additional concept regression loss $\mathcal{L}_{external}$
as follows:
\begin{equation}
    \mathcal{L}_{external}(F) = \mathop{\mathbb{E}}\limits_{\substack{\mathbf{x}}\sim p(\mathbf{x})} \lVert {R \circ F(\mathbf{x})}-\text{Concept}(\mathbf{x})\rVert^2_F,
\label{eqn:external}
\end{equation}
which is described in \Fref{fig:architecture}.
Thereby, the knowledge from the 
external model could be effectively transferred to the image captioning model.
This simple approach 
significantly improves the performance of an image captioning model when the number of paired data is scarce, which is shown in Sections~\ref{sec:web} and \ref{sec:scarce}.

To produce the concept vector, we use the pre-trained relational captioning model~\cite{kim2021dense}.
We generate abundant relational caption proposals from an image by using the model, and each caption is map to an embedding by utilizing Glove word vector~\cite{pennington2014glove}.
Then, in order to represent the global \emph{image-level} concept, we average out all the vectors obtained from the image to 
form a concept vector $\mathbf{v}$ of the image.
The concept vector encodes semantic concept of the scene.

The total loss function for training our model is as follows:
\begin{equation}
\begin{aligned}
    \min_{F,G,H, \{T\}} \max_{D}{ } & \mathcal{L}_{cap}(F,H) + \lambda_1 \tilde{U}(F,G, \{T\},D) \\
& + \lambda_2  \mathcal{L}_{triplet}(F,H) + \lambda_3 \mathcal{L}_{external}(F),
\end{aligned}
\label{eqn:total_loss}
\end{equation}
where $\mathcal{L}_{cap}$ denotes the captioning loss defined in \Eref{eqn:pseudo_cross_entropy}, $\tilde{U}$ the loss for adversarial training defined in \Eref{eqn:gan}, $\mathcal{L}_{triplet}$  the triplet loss defined in \Eref{eqn:triplet}, 
$\mathcal{L}_{external}$ the concept regression loss defined in \Eref{eqn:external},
and $\lambda_1 = \lambda_2 = \lambda_3 = 0.1$.

\begin{figure*}[t]
\begin{center}
   \includegraphics[width=.68\linewidth]{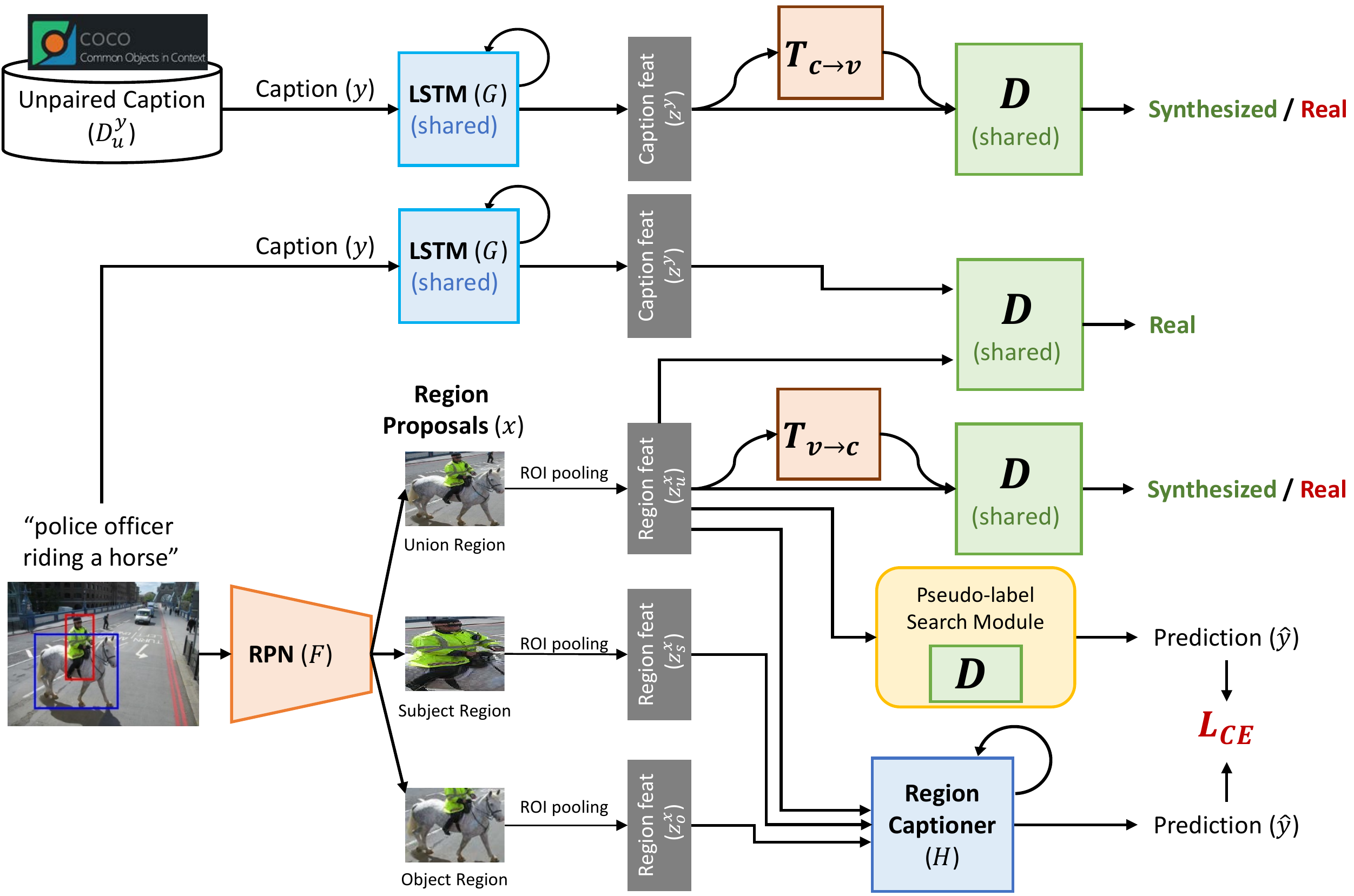}
\end{center}
   \vspace{-2mm}
   \caption{
   Illustration of the proposed semi-supervised \emph{region-based} image captioning structure. 
In addition to the paired region-based image captioning data $\mathcal{D}_p$, we leverage an external image captioning dataset as an unpaired caption dataset $\mathcal{D}^y_u$,
and the instances having no caption label as an unpaired image dataset $\mathcal{D}^x_u$.
   }
   \vspace{-2mm}
   \label{fig:architecture_extended}
\end{figure*}


\subsection{Theoretical Analysis}

In this section, we analyze our minimax-style learning framework and its favorable guarantees, which
include a global equilibrium exists in our learning framework and it is also achievable.
These analyses show that our design of the system and loss functions are 
well-grounded to pursue our objective of the multi-modal distribution match.
To reach this conclusion, we first show the following Lemma~\ref{lemma:optimalD}.

\begin{lemma}
For any fixed generators $F$, $G$, and $\{T\}$, the optimal discriminator $D$ of the minimax game defined by the objective function $U(F, G, \{T\}, D)$ in \Eref{eq:U} is 
\begin{equation}
D^*(\mathbf{z}^x,\mathbf{z}^y) = \frac{p(\mathbf{z}^x,\mathbf{z}^y)}{p(\mathbf{z}^x,\mathbf{z}^y) + p_{1/2}(\mathbf{z}^x,\mathbf{z}^y)},
\end{equation}
where $p_{1/2}(\mathbf{z}^x,\mathbf{z}^y) = \tfrac{(p_{v\rightarrow c}(\mathbf{z}^x,\mathbf{z}^y)+p_{c\rightarrow v}(\mathbf{z}^x,\mathbf{z}^y))}{2}$ is a mixture distribution.
\label{lemma:optimalD}
\end{lemma}

This shows that the optimal discriminator $D^*$ is at the balance between the true data distribution and the mixture distribution defined by $F$, $G$, and $\{T\}$.
Given the fixed $D^*(\mathbf{z}^x,\mathbf{z}^y)$, we can reformulate the minimax game with the function $U(F, G, \{T\}, D)$ as minimizing the sub-problem $V(F, G, \{T\}) = \max_D U$ over $F, G$ and $\{T\}$.
Then, we have the following lemma.

\begin{lemma}
Given $D = D^*(\mathbf{z}^x,\mathbf{z}^y)$, the global minimum of $V(F, G, \{T\})$ is achieved if and only if $p(\mathbf{z}^x,\mathbf{z}^y)=p_{1/2}(\mathbf{z}^x,\mathbf{z}^y)$, and the optimum value is $-\log4$.

Furthermore, the marginal distributions $p(\mathbf{z}^x)$ and~$p(\mathbf{z}^y)$ can be captured by the learned marginal distributions, \ie~$p(\mathbf{z}^y)=\mathop{p}\limits_{c\rightarrow v}(\mathbf{z}^y)=\mathop{p}\limits_{v\rightarrow c}(\mathbf{z}^y)$ and $p(\mathbf{z}^x)=\mathop{p}\limits_{c\rightarrow v}(\mathbf{z}^x)=\mathop{p}\limits_{v\rightarrow c}(\mathbf{z}^x)$.
\label{lemma:optimalG}
\end{lemma}
The standard adversarial training in GAN~\cite{goodfellow2014generative} uses a similar way with Lemma~\ref{lemma:optimalG} and shows that a generator perfectly replicates the data generating process if the optimal discriminator can be found.
However, Lemma~\ref{lemma:optimalG} shows only up to the fact that our model can at least replicate data marginal distributions and a mixture of $\{T\}$ can replicate the joint data distribution.
In the next step, we show that we can actually find a global equilibrium point $p(\mathbf{z}^x,\mathbf{z}^y)= p_{v\rightarrow c}(\mathbf{z}^x,\mathbf{z}^y) = p_{c\rightarrow v}(\mathbf{z}^x,\mathbf{z}^y)$ that mimics the data generating (transformation) process in both directions as follows.



\begin{theorem}
Given an augmented objective function defined as:
\begin{equation}
\begin{aligned}
{U}(F, G, \{T\},D) 
&+ \mathrm{KL}\left[p(\mathbf{z}^x|\mathbf{z}^y)||p_{c\rightarrow v}(\mathbf{z}^x|\mathbf{z}^y)\right]\\
&+ \mathrm{KL}\left[p(\mathbf{z}^y|\mathbf{z}^x)||p_{v\rightarrow c}(\mathbf{z}^y|\mathbf{z}^x)\right].
\end{aligned}
\label{eq:augmented_U}
\end{equation}
The equilibrium of \Eref{eq:augmented_U} is achieved if and only if $p(\mathbf{z}^x,\mathbf{z}^y)= p_{v\rightarrow c}(\mathbf{z}^x,\mathbf{z}^y) = p_{c\rightarrow v}(\mathbf{z}^x,\mathbf{z}^y)$.
\label{theorem:convergence}
\end{theorem}

Lemma~\ref{lemma:optimalG} and Theorem~\ref{theorem:convergence} show that, without the additional regularization, the learned distribution is only matched up to marginal distributions and the true data distribution may be achieved with the non-unique mix of two distributions, $p_{1/2}(\mathbf{z}^x,\mathbf{z}^y)$. 
With the additional regularization, Theorem~\ref{theorem:convergence} shows that the true distribution can be matched with the favorable unique global equilibrium guarantee.
Finally, by Theorem~\ref{theorem:convergence}, we can ensure that $F$, $G$, and $\{T\}$ will converge to the true distribution if $F$, $G$, and $\{T\}$ have enough capacity and each model has been trained to achieve the optimum.

Unfortunately, directly minimizing the KL divergence terms in \Eref{eq:augmented_U} is infeasible in practice.
In \Eref{eqn:gan}, we use the simple alternative of $L_{reg}$ as a practical solution, which can be regarded as a Monte Carlo approximation of distribution matching and is proportional to those matching.
Note that, despite departing from the theoretical guarantees,
the noticeable performance improvement in our empirical study suggests that our method is indeed a reasonable realization of the theory.


\begin{table}[t]
    \centering
    \resizebox{\linewidth}{!}{
    \begin{tabular}{@{}lccc@{}}
    \toprule
        Experiments & Paired data & Unpaired images & Unpaired captions \\
        \midrule
        Partially Labeled  & MS COCO caption & Unlabeled-COCO & \multirow{2}{*}{--} \\
        COCO (\Sref{sec:semi}) & (113k) & (123k) &  \\[1mm]
        {Web-Crawled} &  0.5--1\%
        of MS COCO  & MS COCO    & Shutterstock  \\
        (\Sref{sec:web}) & caption (0.56k to 1.13k)   & caption (112k)  & (2.2M) \\[1mm]
        {Relational Caption} &  Labeled regions  &  Unlabeled regions  & MS COCO caption\\
        (\Sref{sec:relcap}) & of Rel.Cap.   & of Rel.Cap.  &  (113k $\times$ 5) \\[1mm]
        {Scarcely-paired } &  1\% of MS COCO   & 99\% of MS COCO  & 99\% of MS COCO\\
        COCO (\Sref{sec:scarce}) & caption (1.3k)   & caption (112k)  & caption (112k $\times$ 5) \\
    \bottomrule
    \end{tabular}
    }\vspace{1mm}
    \caption{Data source of each experiment setup. The numbers in the parentheses indicate the number of samples. }
    \label{tab:experiment_setups}
\end{table}

\begin{table*}[t]
\centering
    \resizebox{.9\linewidth}{!}{%
		\begin{tabular}{l cccccccccc}\toprule
			&BLEU1	&	BLEU2		&BLEU3	&BLEU4&ROUGE-L&SPICE&METEOR&CIDEr	\\\midrule
			Self-Retrieval~\cite{liu2018show} (w/o unlabeled) 	&		79.8	&	62.3		&	47.1	&34.9 &56.6&20.5&27.5&114.6		\\		
			Self-Retrieval~\cite{liu2018show} (with unlabeled)  &		80.1	&	63.1		&	48.0	&35.8 &57.0 &21.0 &27.4&117.1		\\
			\midrule
			Baseline (w/o unlabeled)  &		77.7	&	61.6  &	46.9&	36.2 &  56.8    &20.0&26.7&114.2		\\
			\textbf{Ours} (w/o unlabeled) &		80.8    &	65.3  &	49.9&	37.6 &  58.7    &   22.7    &   28.4    &   122.6		\\
			\textbf{Ours} (with unlabeled)  &\textbf{81.2}&\textbf{66.0}&\textbf{50.9}&\textbf{39.1}&\textbf{59.4}&\textbf{23.8}&\textbf{29.4}&\textbf{125.5}\\
			\bottomrule
		\end{tabular}
    }
    \vspace{1mm}
    \caption{
	Comparison with the semi-supervised image captioning method, ``Self-Retrieval''~\protect\cite{liu2018show}.
	Our method shows improved performance even without the Unlabeled-COCO data (denoted as \emph{w/o unlabeled}) as well as with the Unlabeled-COCO (\emph{with unlabeled}), although our model is not originally designed for such scenario.
	\vspace{-4mm}
	}
	\label{table:captioning_semi}	
\end{table*}

\begin{table*}[t]
\centering
    \vspace{0mm}
	\vspace{0mm}
    \resizebox{.8\linewidth}{!}{%
		\begin{tabular}{l cccccccccc}\toprule
			&BLEU1	&	BLEU2		&BLEU3	&BLEU4&ROUGE-L&SPICE&METEOR&CIDEr	\\\midrule
			NIC~\cite{vinyals2015show} 	&		72.5	&	55.1		&	40.4		 &	29.4   &52.7&18.2&25.0&95.7		\\
			NIC \textbf{+Ours}      &\textbf{74.2}	&	\textbf{56.9}		&	\textbf{42.0}&	\textbf{31.0}            &\textbf{54.1}&\textbf{19.4}&\textbf{28.8}&\textbf{97.8}		\\
			\midrule
			Att2in~\cite{rennie2017self} 	&		74.5	&	58.1		&	44.0		&	33.1 &54.6 & 19.4&26.2&104.3	\\	
			Att2in \textbf{+Ours}  &\textbf{75.5}	&	\textbf{58.8}		&	\textbf{45.0}		&	\textbf{34.3} &\textbf{55.4}&\textbf{20.0}&\textbf{26.8}&\textbf{105.3}	\\
			\midrule
			Up-Down~\cite{anderson2017bottom} 	&		76.7	&	60.7		&	47.1		&	36.4 &56.6&20.6&27.6&113.6		\\	
			Up-Down \textbf{+Ours}  &		\textbf{77.3}	& \textbf{64.1}		&	\textbf{47.6}		&	\textbf{37.0} &\textbf{57.1}&\textbf{21.0}&\textbf{27.8}&\textbf{114.6}		\\
			\midrule
			
			AoANet~\cite{huang2019attention} 	&		76.5	&	61.5		&	47.1 &35.8 &56.6 &20.8&27.7&116.0		\\	
			AoANet \textbf{+Ours}  &		\textbf{77.0}	& \textbf{62.1}		&	\textbf{48.4}		&	\textbf{37.7} &\textbf{57.0}&\textbf{21.2}&\textbf{28.0}&\textbf{118.1}		\\
			\midrule
			
			$\mathcal{M}^2$ Transformer~\cite{cornia2020meshed}  	            &		77.3	&	61.3		    &	47.8		        &	    37.2        &57.6           &21.2       &28.1           &120.8		\\	
			$\mathcal{M}^2$ Transformer \textbf{+Ours}            &	\textbf{77.7}	& \textbf{61.7}		&	\textbf{48.0}		&	\textbf{37.3} &\textbf{58.0}&\textbf{21.4}&\textbf{28.2}&\textbf{121.2}		\\
			
			\bottomrule
		\end{tabular}
    }
    \vspace{1mm}
    \caption{Evaluation of our method with different backbone architectures as an add-on module. 
	All models are reproduced and trained with the full paired MS COCO caption data and the cross entropy loss.
	Our training method with adding the Unlabeled-COCO images is applied to each method in a semi-supervised way, which shows consistent improvement in all the metrics.
	}
	\vspace{-4mm}
	\label{table:captioning_unpaired}	
\end{table*}


\subsection{Extension to Region-based Captioning}

Our semi-supervised learning method can be extended
to other advanced
 visual captioning tasks. 
In this work, we extend our approach to region-based image captioning tasks, which require to localize object instances in the scenes~\cite{johnson2016densecap,kim2019dense}.
We especially focus on the relational captioning task~\cite{kim2019dense}, where the task is to caption the interactions of object instances in the visual scene, which can be regarded as a generalization of the instance-wise captioning~\cite{johnson2016densecap}.

The pipeline of the relational captioning~\cite{kim2019dense} work is as follows.
Given an input image, $B$ number of object proposals from the region proposal network (RPN)~\cite{ren2015faster} are obtained to localize each object instance.
To take interactions between objects into account, the combination layer~\cite{kim2019dense} produces the subject-object region pairs of the object proposals by 
assigning each instance into either subject or object role, \ie~$B\times(B{-}1)$ subject-object region pairs.
Given a region pair, we obtain a triplet of features consisting of the subject ($\mathbf{z}^x_s$), object ($\mathbf{z}^x_o$), and the union of their regions ($\mathbf{z}^x_u$), as illustrated in \Fref{fig:architecture_extended}.
In this task, our semi-supervised method (illustrated in \Fref{fig:architecture}) is applied to captions in the dataset (denoted as $y$) and 
the union region features (denoted as $\mathbf{z}^x_u$) in addition to the supervised loss with the target task data. 
Thereby, the learned model predicts a large number of relational captions describing each pair of objects in the input image.

As the region-based caption labels in the existing datasets~\cite{johnson2016densecap,kim2019dense} 
are mostly in the form of subject-predicate-object triplet, most descriptive phrases in general can be thought of as following a similar form. 
Therefore, 
we postulate that 
it would be helpful to leverage more natural human-labeled language (caption)
datasets as unpaired caption information $\mathcal{D}^y_u$. 
Also, in these region-based tasks, we can leverage 
the instances 
having no caption label (\ie~negative sample) as an unpaired image dataset $\mathcal{D}^x_u$ as well for further regularization.

%% file: 4_sec_experiment_TMM.tex
In this section, we describe the experimental setups and competing methods and demonstrate the performance of our semi-supervised captioning with both quantitative and qualitative results.

\subsection{Experimental Setups}
\label{setup}

We utilize the MS COCO caption dataset~\cite{lin2014microsoft} (we will refer to MS COCO for simplicity) as our target dataset, which contains $123$k images 
with 5 caption labels per image. 
To validate our model, we follow \emph{Karpathy} splits~\cite{karpathy2015deep}, which have been broadly used in image captioning literature.
The Karpathy splits contain 113k training, 5k validation, and 5k test images in total.
In our experiment, to simulate the scenario that both paired and unpaired data exist, we use four different setups: 
1) partially labeled COCO~\cite{liu2018show},  
2) web-crawled data~\cite{feng2018unsupervised}, 
3) relational captioning data~\cite{kim2019dense},
and 
4) scarcely-paired COCO setup we proposed.
The data source of each experiment setup is described in \Tref{tab:experiment_setups}.



For evaluation, we use the following metrics conventionally used in image captioning: BLEU~\cite{papineni2002bleu}, ROUGUE-L~\cite{lin2004rouge}, SPICE~\cite{anderson2016spice}, METEOR~\cite{denkowski2014meteor}, and CIDEr~\cite{vedantam2015cider}.  
All the evaluation is done on the MS COCO caption test set.

\subsection{Evaluation on Partially Labeled COCO}
\label{sec:semi}

For the \emph{partially labeled} COCO experiment,
we follow Liu~\etal\cite{liu2018show} and use the whole MS COCO caption data (paired) and add the \emph{``Unlabeled-COCO''} split.
The Unlabeled-COCO split includes unpaired images from
the official MS COCO dataset~\cite{lin2014microsoft}, which involves 123k images without any caption label (no additional unpaired caption is used).
Note that the MS COCO caption dataset and the Unlabeled-COCO split do not have overlapped data. 
In this setup, a separate \emph{unpaired} caption data $\mathcal{D}_u^y$ does not exist. 
To compute the cross entropy loss, 
we apply the pseudo-label assignment to the Unlabeled-COCO images.
We use captions from the paired COCO data $\mathcal{D}_p$ as pseudo-label candidates.

We compare with the recent semi-supervised image captioning method, called \emph{Self-Retrieval} \cite{liu2018show}, on the partially labeled COCO setup in \Tref{table:captioning_semi}. 
For a fair comparison with it, we replace the cross entropy loss from our loss with the policy gradient method~\cite{rennie2017self} to directly optimize {our} model with the CIDEr score as in \emph{Self-Retrieval}~\cite{liu2018show}.
As our baseline model (denoted as \emph{Baseline}), we train a model only with the policy gradient method without the proposed GAN model.
The results show that, when only using the 100\% paired MS COCO caption dataset (denoted as \emph{w/o unlabeled}), our model already shows improved performance over Self-Retrieval.
Moreover, when adding the Unlabeled-COCO images (denoted as \emph{with unlabeled}), our model outperforms Self-Retrieval in all the metrics even without the concept transfer method.
The results suggest that our method is advantageous in the semi-supervised setup.


To further validate our method in the semi-supervised setup, 
we compare on different advanced backbone architectures equipped with attention mechanism~\cite{anderson2017bottom,rennie2017self,vinyals2015show}, self-attention approach~\cite{huang2019attention}, and the recent Transformer based architecture~\cite{cornia2020meshed}, 
which were originally
developed for fully-supervised methods.
We use the same data setup with the above, but we replace CNN ($F$) and LSTM ($H$) in our framework with the image encoder and the caption decoder of their image captioning models. 
Then, these models are trained by our learning method as it is without the concept transfer method, which consists of alternating between the discriminator update and pseudo-labeling.
\Tref{table:captioning_unpaired} shows that training with the additional Unlabeled-COCO data via \emph{our} training scheme consistently improves all the baselines in all the metrics.



\begin{table}[t]
\centering
    \vspace{0mm}
    \resizebox{1.0\linewidth}{!}{%
		\begin{tabular}{l ccccc}\toprule
			&		BLEU4&ROUGE-L&SPICE&METEOR&CIDEr	\\\midrule
			Paired only (0.5\% paired)	  &   4.4	&33.7   &3.6    &10.8   &8.6		\\
			\textbf{Ours} (0.5\%)  &5.4      &34.6   &4.2    &12.0   &10.5		\\
			\textbf{Concept} (0.5\%)	  &   11.7	&40.7   &7.0    &14.2   &27.0		\\
			\textbf{Ours + Concept} (0.5\%)  &\textbf{12.2} ($\times$2.3)      &\textbf{41.6} ($\times$1.2)  &\textbf{7.6} ($\times$1.8)   &\textbf{14.8} ($\times$1.2)  &\textbf{30.5}	($\times$2.9)	\\
			\midrule
			Paired only (0.7\% paired)	  &   3.5	 &36.1  &3.7    &11.4   &8.9		\\
			\textbf{Ours} (0.7\%)        &8.5      &39.0   &5.2    &13.6   &20.2		\\
			\textbf{Concept} (0.7\%)	    &   13.3	&42.0   &8.5    &15.4   &33.6		\\
			\textbf{Ours + Concept} (0.7\%)  &\textbf{14.3} ($\times$1.7)      &\textbf{42.8} ($\times$1.1)  &\textbf{9.2} ($\times$1.8)    &\textbf{16.3} ($\times$1.2)   &\textbf{38.9} ($\times$1.9)		\\
			\midrule
			Paired only (0.8\% paired)	  &   8.8	 &39.1  &5.9    &13.2   &21.9		\\
			\textbf{Ours} (0.8\%)  &   12.2	 &41.6  &7.6    &15.1   &29.0	\\
			\textbf{Concept} (0.8\%)	  &   14.8	&43.2   &9.4    &16.4   &39.5		\\
			\textbf{Ours + Concept} (0.8\%)  &   \textbf{15.5} ($\times$1.3)	 &\textbf{44.0} ($\times$1.1)  &\textbf{10.2} ($\times$1.3)    &\textbf{16.9} ($\times$1.1)   &\textbf{42.0} ($\times$1.4)	\\
			\midrule
			Paired only (1\% paired)	 &   13.4   &41.9   &8.3    &15.9   &36.0		\\
			\textbf{Ours}  (1\%)  &15.2      &43.3   &9.4    &16.9   &39.7		\\
			\textbf{Concept} (1\%)	  &   16.2	&   44.0 &   10.1 &17.2   &44.5		\\
			\textbf{Ours + Concept}  (1\%)  &\textbf{17.4} ($\times$1.1)      &\textbf{45.0} ($\times$1.04)   &\textbf{10.9} ($\times$1.2)    &\textbf{17.9} ($\times$1.1)   &\textbf{47.7} ($\times$1.2)		\\
			\midrule
			Feng~\etal\cite{feng2018unsupervised} &5.6        &28.7   &8.1    &12.4   &28.6\\
			Guo~\etal\cite{guo2020recurrent} & 6.4        &31.3   &9.1    &13.0   &29.0\\
			Zhu~\etal\cite{zhu2022unpaired} & 5.9        &28.0   &7.6    &12.0   &26.9\\
			\bottomrule
		\end{tabular}
    }
    \vspace{1mm}
    \caption{Performance comparison with web-crawled data (Shutterstock).
    On top of unpaired image and caption data, our method is trained with 0.5 -- 1\% of paired data, while Feng~\etal and Guo~\etal use 36M additional images of the OpenImage dataset.
    We also show the relative performance improvements before and after applying the concept transfer on our method, and add the comparison with a baseline ``Concept,'' which only uses paired only data with the concept transfer, for reference.
    Applying the concept transfer significantly improves the image captioning performance, especially when the number of paired samples is scarce.
	\vspace{-8mm}
	}
	\label{table:shutterstock}	
\end{table}

\subsection{Evaluation on Web-Crawled Data Setup}
\label{sec:web}
To simulate a more realistic scenario involving crawled data from the web, we use the setup suggested by Feng~\etal\cite{feng2018unsupervised}.
They collect a sentence corpus by crawling the image descriptions from Shutterstock\footnote{https://www.shutterstock.com} as unpaired caption data $\mathcal{D}_u^y$, whereby 2.2M sentences are collected.
For unpaired image data $\mathcal{D}_u^x$, they use only the images from the MS COCO data, 
while the captions are not used for training.
For training our method, we leverage from 0.5\% to 1\% of the paired MS COCO caption data as our paired dataset $\mathcal{D}_p$, \ie~very scarce data with a few hundreds or a thousand.
This is an extremely challenging scenario as the paired and unpaired datasets are disjoint with different domains. In other words, there is no guarantee that all unpaired samples have their exact matches in the counterpart dataset.
The results are shown in \Tref{table:shutterstock} including the comparison with
Feng~\etal, Guo~\etal\cite{guo2020recurrent}, and Zhu~\etal\cite{zhu2022unpaired}.
Note that all of Feng~\etal, Guo~\etal, and Zhu~\etal exploit external large-scale data, \ie~36M images of the OpenImages dataset.
Up to 0.7\% of paired only data (793 pairs), the baseline shows lower scores in terms of BLEU4 and METEOR than Feng~\etal, while Ours shows comparable or favorable performance in BLEU4, ROUGE-L and METEOR against Feng~\etal, Guo~\etal, and Zhu~\etal.
Ours starts to have significantly higher scores in all the metrics from 1\% of paired data (1,133 pairs), even without external knowledge.

Moreover, additionally applying the concept transfer with additional loss in \Eref{eqn:external}
by exploiting relational captions~\cite{kim2019dense}
(denoted as Ours $+$ Concept)
shows significant performance improvement,
especially when the number of paired samples is scarce.
Note that although applying the concept transfer to the Paired only baseline also shows noticeable performance improvement, combining both Ours and the concept transfer consistently shows the best performance in all settings.
With 0.5\% paired data, compared to our model without the concept transfer (Ours), our final model (Ours $+$ Concept) shows near 2 times performance improvement on average; in particular, almost 3 times in terms of the CIDEr metric.

\begin{table}
\vspace{0mm}
    \centering
    \resizebox{.9\linewidth}{!}{%
         \begin{tabular}{l ccc}\toprule
										&	mAP (\%) 	&Img-Lv. Recall	&METEOR\\\midrule
            \texttt{Direct Union}		&		--	&	17.32	&11.02\\\midrule
           \texttt{{MTTSNet}}~\cite{kim2019dense} &{0.88}&{34.27}	&{18.73}\\
            \texttt{{MTTSNet}$^\dagger$}~\cite{kim2021dense}&{1.12}&{45.96}	&{18.44}\\
            \midrule
            \texttt{{MTTSNet}$^\dagger$ + \textbf{Ours}}&\textbf{1.19}&\textbf{47.25}&\textbf{19.03}\\
            \bottomrule
		\end{tabular}
        }
        \vspace{1mm}
	\caption{Evaluation on the relational dense captioning result with the Relational Captioning dataset~\cite{kim2019dense}. 
	We annotate the extended MTTSNet (MTTSNet + Relational module) by Kim~\etal\cite{kim2021dense} with $\dagger$.
	The extended MTTSNet trained with the proposed framework shows improvement over the one without the proposed framework by a noticeable margin.}
	\vspace{-4mm}
	\label{table:relcap}	
\end{table}

\subsection{Evaluation on Relational Captioning Task}
\label{sec:relcap}
We apply our semi-supervised learning method to a dense relational object region based image captioning task, \ie~relational captioning~\cite{kim2019dense}.
For evaluation, we use the Relational Captioning dataset~\cite{kim2019dense} consists of 85,200 images with 75,456 / 4,871 / 4,873 splits for train / validation / test sets, respectively.
We regard the whole paired Relational Captioning dataset as our paired data $\mathcal{D}_p$, and we utilize the captions from the MS COCO caption dataset as the unpaired caption dataset $\mathcal{D}^y_u$.
In particular, we define the visual features in the training batch ($\mathbf{z}^x$) as the region features from individual object regions.
As the relational captioning is a region based task, we utilize the negative regions with no captions label as the unpaired image dataset $\mathcal{D}^x_u$.
We apply our method to the extended version of MTTSNet (MTTSNet + Relational embedding module; annotated with $\dagger$) by Kim~\etal\cite{kim2021dense} and compare with the other strong baselines.

\begin{table}[t]
\centering
    \resizebox{1\linewidth}{!}{%
		\begin{tabular}{l cc cc}\toprule
													&		Recall	&	METEOR		&\#Caption	&Caption/Box	\\\midrule
			Relational Cap. (\texttt{Union})	&		38.88	&	18.22	&	85.84	&	9.18	\\
			Relational Cap. (\texttt{{MTTSNet}}~\cite{kim2019dense})		&{46.78} &{21.87}	&{89.32}&{9.36}\\
			Relational Cap. (\texttt{{MTTSNet}$^\dagger$}~\cite{kim2021dense})		&{56.52} &{22.03}	&{80.95}&{9.24}\\
			Relational Cap. (\texttt{{MTTSNet}$^\dagger$} + \textbf{Ours})		&\textbf{61.40} &\textbf{23.88}	&{89.46}&{9.65}\\\bottomrule
		\end{tabular}
    }
    \vspace{0mm}
	\caption{Comparisons of the holistic image captioning on the Relational Captioning dataset~\cite{kim2019dense}. 
	\vspace{-4mm}
	}
	\label{table:recall}	
\end{table}

\begin{figure*}
\centering
   \includegraphics[width=0.85\linewidth]{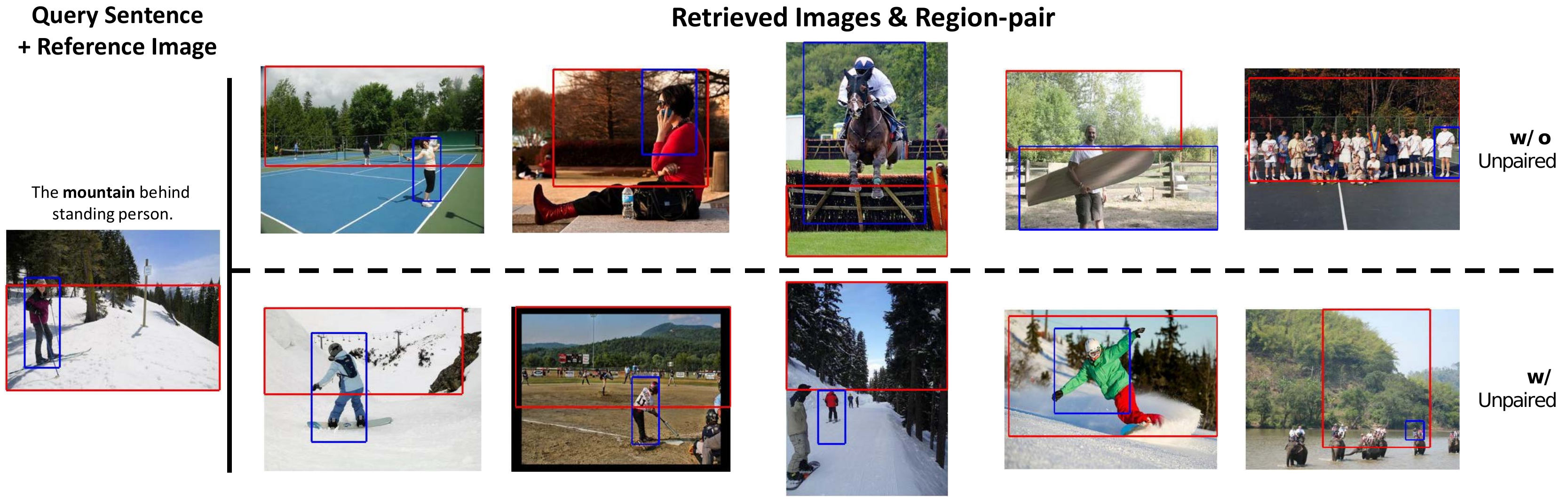}\\
   \includegraphics[width=0.85\linewidth]{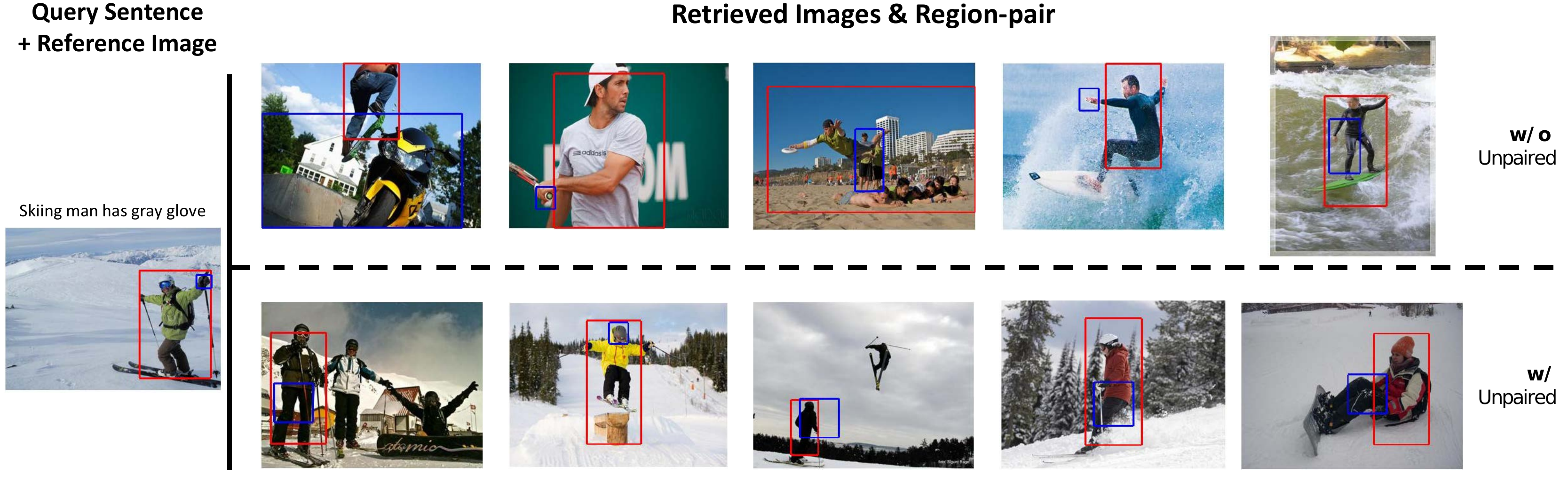}\\
   \caption{Qualitative results of the caption-based image retrieval on the Relational Captioning dataset~\cite{kim2019dense}. 
   The results are obtained by the relational captioning methods, which improves the caption-based image retrieval in multiple aspects.
   MTTSNet without the proposed framework (w/o Unpaired) shows a few incorrect retrieval results, whereas the extended MTTSNet trained with our framework (w/ Unpaired) correctly retrieves image region-pairs.}
   \vspace{-2mm}
\label{fig:retrieval_RelCap}
\end{figure*}

We follow the evaluation protocols suggested by Kim~\etal\cite{kim2019dense}.
The relational dense captioning performance on the Relational Captioning dataset is shown in Tables~\ref{table:relcap} and \ref{table:recall}.
In addition, the relational dense captioning performance on the VRD dataset~\cite{lu2016visual} is shown in \Tref{table:ablation_VRD}.
The extended MTTSNet trained with our proposed method shows improvement
by a noticeable margin over the MTTSNet counterpart in all the metrics and all the tables.

We also show the the caption based image region-pair retrieval results in \Fref{fig:retrieval_RelCap} as an application.
As the Relational Captioning dataset might have limited generalizability, 
MTTSNet without the proposed framework (denoted as w/o Unpaired) shows several incorrect retrieval results, whereas the extended MTTSNet trained with our framework (denoted as w/ Unpaired) correctly retrieves image region-pairs.
Note that, even if the MTTSNet without our framework retrieves correct images, the semantic reasoning in the region-pairs is incorrect when we do not leverage external knowledge.
We also show the quantitative results of the retrieval in \Tref{table:retrieval}.
Similar to the other experiment, the extended MTTSNet with our framework shows favorable image retrieval performance in all the metrics, which demonstrates our method is beneficial to the application level as well.

\begin{table}[t]
\resizebox{1.0\linewidth}{!}{%
\begin{tabular}{l ccc}\toprule
&	mAP (\%)	&	Img-Lv. Recall 	& METEOR\\\midrule

\texttt{Direct Union}&	--	&	54.51	&25.53\\\midrule
\texttt{{MTTSNet}}~\cite{kim2019dense}			&{2.18}& 	{71.44}	&{35.47}\\
\texttt{{MTTSNet$^\dagger$}}~\cite{kim2021dense}			&{2.21}& 	{73.36}	&{35.65}\\
    {\texttt{MTTSNet}$^\dagger$ + \textbf{Ours}}			&\textbf{2.58}& 	\textbf{73.47}	&\textbf{35.97}\\
\midrule
\texttt{Language Prior}~\cite{lu2016visual} &2.13 & 46.60&28.12 \\
\texttt{Shuffle-Then-Assemble}~\cite{yang2018shuffle}& 2.20 & 69.98&29.50 \\
\bottomrule
\end{tabular}
}
\vspace{1mm}
\caption{{Evaluation on the relational dense captioning task with the VRD dataset~\cite{lu2016visual}. 
The extended MTTSNet trained with our method shows the best performance among the baselines and the competing methods.
} 
\vspace{0mm} 
}
\vspace{-4mm}
\label{table:ablation_VRD}
\end{table}

\subsection{Analysis on Scarcely-paired COCO}
\label{sec:scarce}
In order to understand the algorithmic  characteristic of our method,
we also provide extensive and comprehensive analysis on our scarcely-paired COCO dataset.
For the \emph{scarcely-paired} COCO setup, we remove the pairing information of the MS COCO caption dataset, while leaving a small fraction of pairs unaltered.
We randomly select only $1\%$ 
of the total data as the paired training data $\mathcal{D}_p$, and remove the pairing information of the rest to obtain unpaired data $\mathcal{D}_u$. 
This dataset allows to evaluate the proposed framework by assessing whether small paired data can lead to learn plausible pseudo-label assignment, and what performance can be achieved compared to the fully supervised case.
We follow the same setting with Vinyals~\etal\cite{vinyals2015show}, if not mentioned. 
The performance evaluated on the MS COCO caption test set is reported.

\begin{table}[t]
	\centering 
    \resizebox{1\linewidth}{!}{%
    \begin{tabular}{l cccc}\toprule
    &		R@1		&	R@5	&	R@10		& 	Med	\\\midrule
    Image Cap. (\texttt{Full Image RNN})~\cite{karpathy2015deep}	    &		9	&	27	&	36	&	14	\\
    Dense Cap. (\texttt{DenseCap})~\cite{johnson2016densecap}		    &		25	&	48	&	61	&	6	\\
    Dense Cap. (\texttt{TLSTM})~\cite{Yang_2017_CVPR}		&		27	&	52	&	67    &	5	\\
    \midrule
    {\texttt{MTTSNet}}~\cite{kim2019dense}		&{29} &{60}	&{73}&{4}\\
    {\texttt{MTTSNet}$^\dagger$}~\cite{kim2021dense}	&{32} &{64}	&{79}&\textbf{3}\\
    {\texttt{MTTSNet}$^\dagger$ + \textbf{Ours}}		&\textbf{34} &\textbf{66}	&\textbf{82}&\textbf{3}\\
    \midrule
    Random chance & 0.1 & 0.5	& 1.0 & - \\\bottomrule
    \end{tabular}    }
	\vspace{1mm}
	\caption{Caption-based image retrieval results on the Relational Captioning dataset~\cite{kim2019dense}. Our framework improves the performance of the MTTSNet even in the image retrieval application across all the metrics.}
	\vspace{-4mm}
    \label{table:retrieval}	
\end{table}

\begin{table}[t]
\centering
    \resizebox{1.0\linewidth}{!}{%
		\begin{tabular}{l|| ccc|cccc}\toprule
			&		(A)	&(B)&	(C)		&BLEU1&BLEU4&METEOR&CIDEr 	\\\midrule 
			Fully paired (100\%)  &		 	&	 		&	 		&72.5 &	29.4 &25.0&95.7		 \\ 
			\midrule
			Paired only (1\%)	                &		     &	          &		      &58.1 &   13.4	 &15.9&36.0		\\ 
			Zhu~\etal\cite{zhu2017unpaired} &            &	          &		       &58.7&	14.1 &16.2&37.7		\\
			\textbf{Ours ver1}              &$\checkmark$&            &		       &60.1&	15.7 &16.5&45.3		\\
			\textbf{Ours ver2}              &$\checkmark$&$\checkmark$&		       &61.3&	17.1 &19.1&51.7		\\
			\textbf{Ours}           &$\checkmark$&$\checkmark$&$\checkmark$&\textbf{63.0}&\textbf{18.7} &\textbf{20.7}&\textbf{55.2}		\\
			\midrule
			Gu~\etal\cite{gu2018unpaired} &&& &46.2& 5.4 &   13.2    &   17.7  \\ 
		    Feng~\etal\cite{feng2018unsupervised} &&& &58.9& 18.6 &   17.9    &   54.9  \\ 
		    Lania~\etal\cite{laina2019towards} &&& & -- & 19.3 &   20.2    &   61.8  \\ 
		    Gu~\etal\cite{gu2019unpaired} &&& &67.1& 21.5 &   20.9    &   69.5  \\ 
		    Chen~\etal\cite{chen2021self} &&& &64.5& 22.5 &   20.0    &   62.4 \\ 
		    Zhu~\etal\cite{zhu2022unpaired} &&& &-& 21.5 &20.1    &   65.7 \\ 
		    \midrule
			\textbf{Ours + Concept}           &$\checkmark$&$\checkmark$&$\checkmark$&\textbf{67.5}&
			\textbf{23.0} &\textbf{23.1}&\textbf{70.1} \\ 
			Paired only + \textbf{Concept}           &&&&61.1&17.3 &17.8&48.0 \\ 
			\bottomrule
		\end{tabular}
    }
    \vspace{1mm}
    \caption{Captioning performance comparison on the MS COCO caption test set.
	The ``Paired only'' baseline is trained only with 1\% of paired data from our \emph{scarcely-paired COCO} dataset, while CycleGAN and Ours \{ver1, ver2, final\} are trained with our \emph{scarcely-paired COCO} dataset (1\% of paired data and unpaired image and caption datasets).
	We denote the ablation study as: (A) the usage of the proposed GAN that distinguishes real or fake image-caption pairs, (B) pseudo-labeling, and (C) noise handling by sample re-weighting.
    In addition, we extend Ours (final) by adding the concept transfer and compare with a baseline ``Paired only + Concept'' for reference.
    We also compare with Gu~\etal\cite{gu2018unpaired}, Feng~\etal\cite{feng2018unsupervised}, Lania~\etal\cite{laina2019towards}, Gu~\etal\cite{gu2019unpaired}, and Chen~\etal\cite{chen2021self} which are trained with unpaired datasets.
	\vspace{-4mm}
	}
	\label{table:ablation}	
\end{table}


In \Tref{table:ablation}, we compare our method with several baselines: 
\emph{Paired Only}; we train our model only on the small fraction (1\%) of the paired data, \emph{CycleGAN}; 
we train our model with the cycle-consistency loss~\cite{zhu2017unpaired}. 
Additionally, we train variants of our model denoted as \emph{Ours} (ver1, ver2, and final). 
\emph{Ours ver1} is the base model trained with our GAN model (\Eref{eqn:gan}) that distinguishes real or fake image-caption pairs.
Even without pseudo-labeling, GAN training unpaired image and caption data already helps better training the encoder networks in an unsupervised way, which improves the image captioning performance.
As one could expect, semi-supervising with unpaired samples from MS COCO data is more helpful to improve the performance than with unpaired web-crawled samples in \Tref{table:shutterstock}.
\emph{Ours ver2}  
adds training with pseudo-labeled unpaired data using 
\Eref{eqn:pseudo_cross_entropy} to \emph{Ours ver1}, while setting the confidence scores $\alpha^x{=}\alpha^y{=}1$ for all training samples.
\emph{Ours (final)} add 
the noise handling technique to \emph{Ours ver2}, which 
is done by re-weighting each sample in the loss (\Eref{eqn:pseudo_cross_entropy}) with the confidence scores {$\alpha^x$ and $\alpha^y$}.
We present the accuracy of the fully supervised (\emph{Fully paired}) model using 100\% of the MS COCO caption training data for reference.

As shown in \Tref{table:ablation}, in a scarce data regime, utilizing the unpaired data improves the captioning performance in terms of all metrics by noticeable margins.
Also, our models show favorable performance compared to the CycleGAN model in all the metrics.
Our final model with the pseudo-labels and the noise handling achieves the best performance in all metrics 
among the baselines.
In addition, applying our concept transfer by utilizing relational captions~\cite{kim2019dense} as an external knowledge (Ours+Concept)
further 
improves the image captioning performance with noticeable margins.
Note that the CIDEr score of our final model with the concept transfer is almost 2
times that of the Paired only baseline.
Also, note that applying our concept transfer on the Paired only baseline shows lower improvement than that of Ours, indicating that the concept transfer is helpful when combined with our semi-supervised learning framework and our integration is non-trivial.

We also compare the recent unpaired image captioning methods~\cite{chen2021self,feng2018unsupervised,gu2018unpaired,gu2019unpaired,laina2019towards,zhu2022unpaired} in  \Tref{table:ablation}.
In 
Gu~\etal\cite{gu2018unpaired}, the AIC-ICC image-to-Chinese dataset~\cite{wu2017ai} is used as unpaired images $\mathcal{D}_u^x$ and the captions from the MS COCO caption dataset are used as unpaired captions $\mathcal{D}_u^y$.
Note that our dataset setup is unfavorable to our method; in that
Gu~\etal\cite{gu2018unpaired} use a far larger amount of additional labeled data (10M Chinese-English parallel sentences of the AIC-MT dataset~\cite{wu2017ai}), Feng~\etal and Laina~\etal\cite{laina2019towards} use 36M samples of the additional OpenImages dataset, and Gu~\etal\cite{gu2019unpaired} use scene graphs from the Visual Genome dataset~\cite{krishna2017visual} (108k).
In contrast, our model only uses a small amount of paired samples (1k) and 122k unpaired data.
Despite far lower reliance on paired data, 
our final model shows favorable performance against the recent unpaired image captioners in all the metrics.



\begin{figure}[t]
\vspace{0mm}
\centering
   \includegraphics[width=0.6\linewidth]{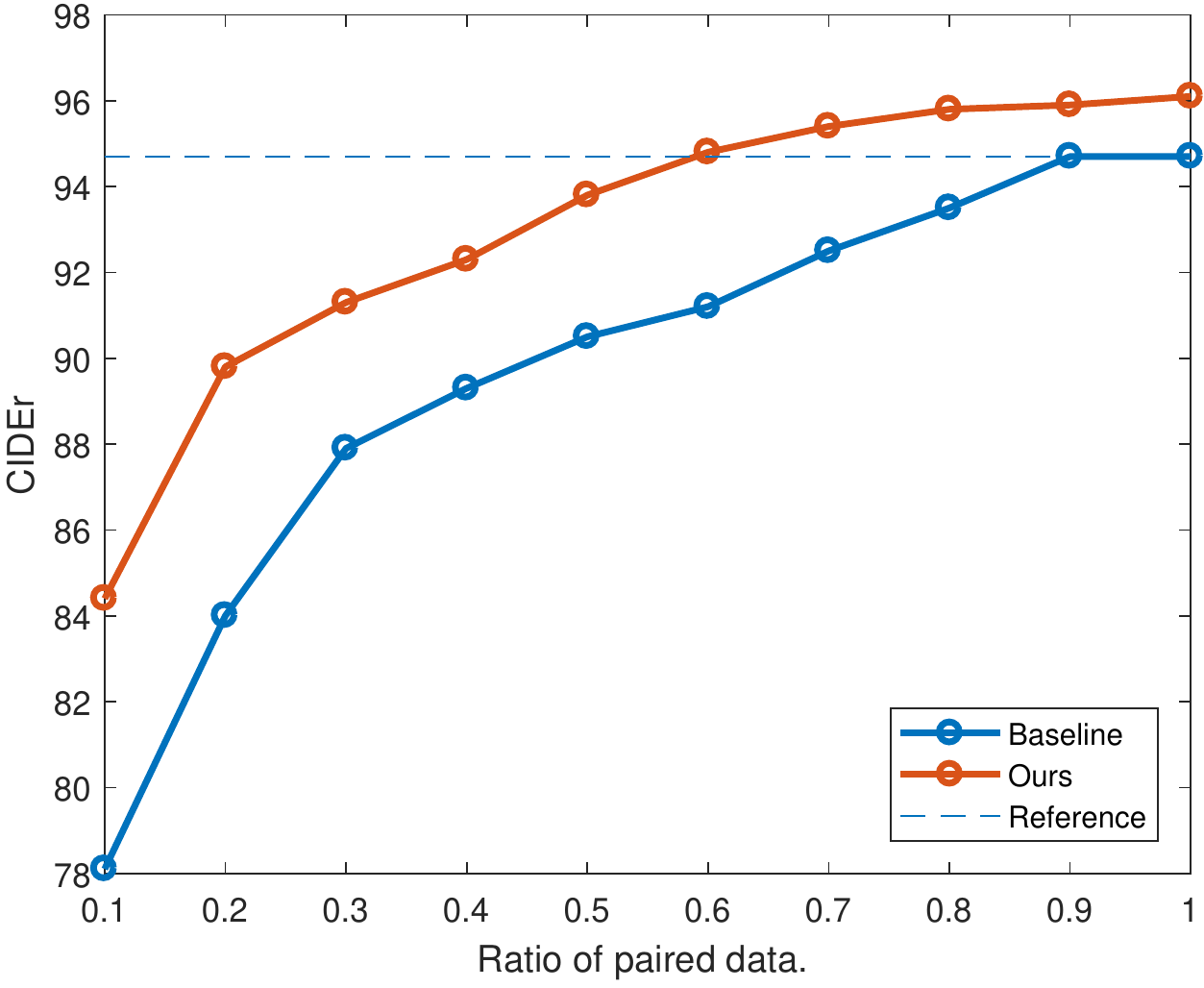}
    \vspace{-2mm}
   \caption{Performance w.r.t. the amount of paired data for training. 
   ``Baseline'' denotes our \emph{Paired Only} baseline, ``Ours'' is our final model, and ``Reference'' is \emph{Paired Only} trained with the full paired data.
   \vspace{-2mm}}
\label{fig:paired}
\end{figure}

\begin{figure*}[!ht]
\centering
   \includegraphics[width=1.0\linewidth]{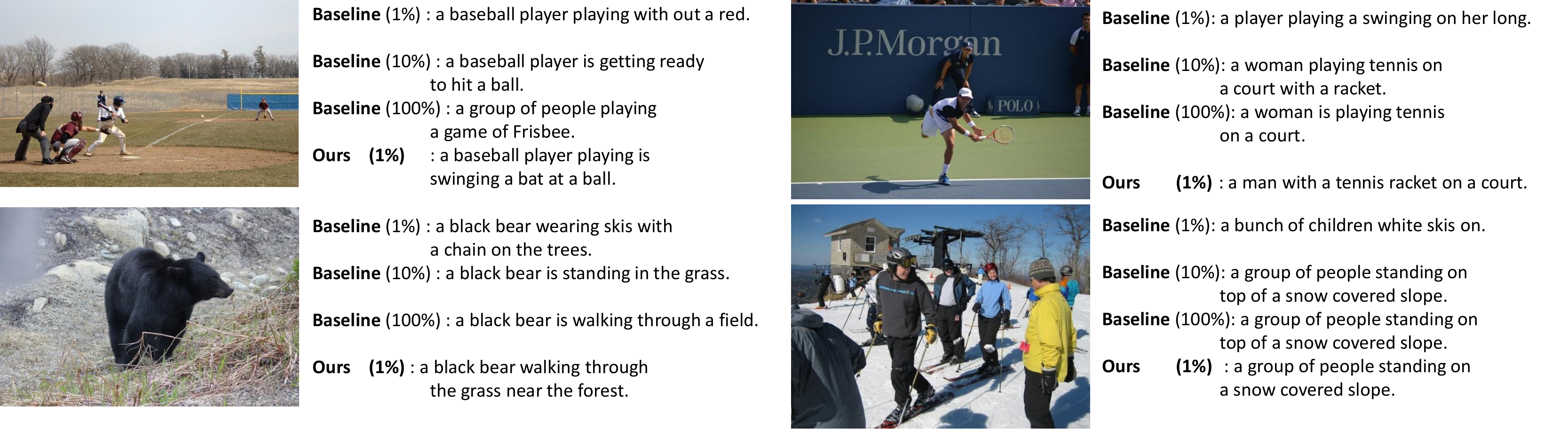}
    \vspace{-7mm}
   \caption{Sampled qualitative results of our model.
   We compare with the baseline models trained only on $N\%$ of paired samples out of the full MS COCO caption dataset.
   Despite the use of only 1\% paired data, our model
   generates reasonable captions comparable to that of the baseline models trained with more data (10\% and above).   }
   \vspace{-4mm}
\label{fig:qualitative}
\end{figure*}

Next, we study the effects of other ratios of paired data used for training 
(in Figs.~\ref{fig:paired}~and~\ref{fig:qualitative}).
We study our final model against our \emph{Paired Only} baseline
according to varying amounts of paired training data
{in \Fref{fig:paired}}, so that we can see 
how much information can be gained from the unpaired data similar to the active learning works~\cite{cho2022mcdal,kim2021single,shin2021labor}.
From 100\% to 10\%, as the amount of paired samples decreases, the fluency and the accuracy of description get worse.
In particular, we observe that most of the captions generated from the {\emph{Paired Only} baseline} trained with 10\% of paired data (11,329 pairs) show erroneous grammatical structures.
In contrast, by leveraging unpaired data, our method generates more fluent and accurate captions, compared to \emph{Paired Only}  trained on the same amount of paired data. 
Note that
our model trained with 60\% of paired data (67,972 pairs) achieves similar performance to the \emph{Paired Only} baseline trained with fully paired data (113,287 pairs) already.
This signifies that our method can save \emph{near half} of the human labeling effort used to construct a dataset.

In \Fref{fig:qualitative}, the \emph{Paired Only} baseline trained with 1\% paired data produces erroneous captions, and the baseline with 10\% paired data 
starts to produce plausible captions.
It is based on $10\times$ more number of paired samples, compared to our model that uses only 1\% of them.
We highlight that, in the two examples on the top row {of \Fref{fig:qualitative}}, our model generates more accurate captions than the {\emph{Paired Only}} baseline trained on the 100\% paired data (``baseball'' to ``Frisbee'' on the top-left, and ``man'' to ``woman'' on the top-right).
This suggests that unpaired data with our method effectively boosts the performance
especially when paired data is scarce.

\begin{figure}[t]
\centering
   \includegraphics[width=1.0\linewidth]{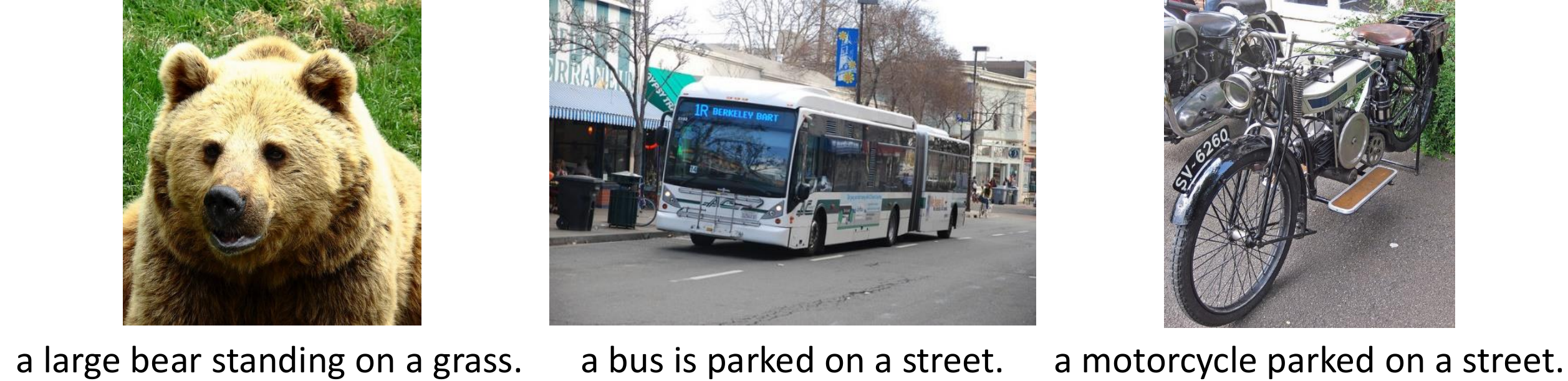}
   \vspace{-6mm}
   \caption{Examples of the pseudo-labels (captions) assigned to the unpaired images. Our model is able to sufficiently and plausibly assign image-caption pairs through the proposed adversarial training.}
\label{fig:pseudo-label}
\end{figure}

In order to demonstrate the effectiveness
of our pseudo-label assignment,
we show the pseudo-labels (captions) assigned to unlabeled images from Unlabeled-COCO images in \Fref{fig:pseudo-label}.
As shown in the figure, despite not knowing the real pairs of these images, the reasonable pseudo-labels are 
assigned by our model.
Note that even though there is no ground truth caption for unlabeled images in the searching pool, the model can find the most likely (semantically correlated) image-caption pair for the given images.

\begin{figure}[t]
\centering
   \includegraphics[width=1\linewidth]{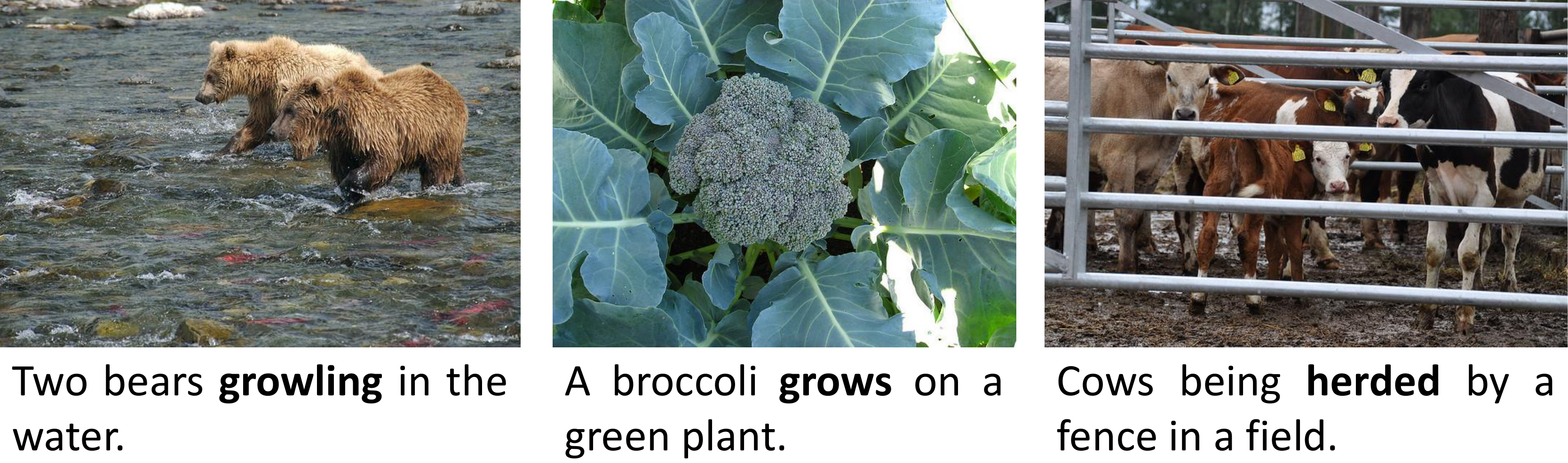}
\vspace{-6mm}
   \caption{Generated caption samples containing words that do not exist in the paired dataset $\mathcal{D}_p$.
   The novel words that are not in $\mathcal{D}_p$ but in $\mathcal{D}_u^y$ are highlighted in bold.}
   \vspace{-2mm}
\label{fig:novel}
\end{figure}

\Fref{fig:novel} highlights interesting results of our caption generation, where the results contain words that do not exist in the paired data of the scarcely-paired COCO dataset.
It shows that the pseudo-label assignment during training enables to properly learn the semantic meaning of the words, such as ``growling,'' ``growing,'' and ``herded,'' which exist in the unpaired caption dataset but not in the paired dataset.
These examples suggest that our method is capable to infer the semantic meaning of  unpaired words to some extent, which would have been unable to be learned with only little paired data. 
This evidences that our method is capable to align abstract semantic spaces between two modalities, \ie~visual data and text.

\noindent\textbf{Scarcely-Paired Setup with Different Domains.}
Training with the unpaired data from the MS COCO dataset would be different from the unpaired data retrieved from the web.
Thus, we also test with a similar but different domain data from the MS COCO dataset.
We use the Flickr30k dataset~\cite{young2014image} as another unpaired image-caption dataset to see whether a small amount of paired information allows to learn matches among
unpaired data in a different domain as well. 
We add images from the Flickr30k dataset, denoted as Flickr, and also add the complement set of the 1\% paired data of the scarcely-paired COCO, denoted as Unpaired.
\Tref{table:ablation_Flickr} still shows consistent improvement even if the MS COCO captions and the independent corpus are mixed.
Given all the results, we would like to note that totally randomly sampled 1\% of paired data (very scarce) would be very challenging to make other models learn to match any general image-caption pair out of the entire set (Unpaired-COCO + Flickr images). 
In this sense, our improvements shown in this experiment are non-trivial.

\begin{table}[t]
\centering
    \resizebox{1\linewidth}{!}{%
		\begin{tabular}{l cccc}\toprule
			&		BLEU1&BLEU4&METEOR&CIDEr	\\\midrule
			Paired only (1\%)&  58.1 &13.4	 &15.9&36.0		\\
			Ours (1\%+Flickr)  &     59.8       &14.7 &16.4&39.2		\\
			Ours (1\%+Unpaired)  &       63.0     &18.7 &20.7&55.2		\\
			Ours (1\%+Unpaired+Flickr)    &\textbf{64.4} &	\textbf{19.2} &\textbf{21.2}&\textbf{55.5}		\\
			\bottomrule
		\end{tabular}
    }
    \vspace{0mm}
    \caption{Dataset ablation study on the Flickr30k dataset with the scarcely-paired COCO setup. 
	\vspace{-4mm}
	}
	\label{table:ablation_Flickr}	
\end{table}

\noindent\textbf{Scarcely-Paired Setup with Unlabeled-COCO Images.}
Both images and captions in the scarcely-paired data come from the original MS COCO caption dataset.
While it is still suitable for analyzing and validating the evaluation, it would have a gap from unpaired data in the wild, because we can find the right caption within the list of unpaired captions in this case.
The right caption for an image may not even be available from the list of unpaired captions in practice. 
To simulate such practical scenarios with real unpaired data, we run a test with a similar setup with \Tref{table:ablation},
but by replacing unpaired images $\mathcal{D}_u^x$ of the scarcely-paired COCO dataset to Unlabeled-COCO images, so that there are no inherent matches between unpaired images $\mathcal{D}_u^x$ and captions $\mathcal{D}_u^y$. 
The results are presented in \Tref{table:ablation_semi} (refer to \Tref{table:ablation} for the methods).
The table shows the same trend as those in \Tref{table:ablation}, which shows that our method is generalizable to the practical scenarios.

\begin{table}[t]
\centering
	\vspace{0mm}
    \resizebox{0.85\linewidth}{!}{%
		\begin{tabular}{l cccc}\toprule
			&		BLEU1&BLEU4&METEOR&CIDEr	\\\midrule
			Paired only (1\%)& 58.1         & 13.4	        &15.9           &36.0		\\
			CycleGAN        &58.5           & 14.0          & 16.1          & 37.1		\\
			Ours ver1       & 59.8          & 15.7          & 16.4          & 39.2\\
			Ours ver2       & 60.9          & 15.8          & 18.6          & 43.3\\
			Ours (final)    & \textbf{61.9} & \textbf{17.7} & \textbf{19.7} & \textbf{49.4}\\
			\bottomrule
		\end{tabular}
    }
    \vspace{1mm}
    \caption{Evaluation on the scarcely paired setup with unpaired images from the different dataset. We train our method by replacing the unpaired images in our scarcely-paired COCO setup to the Unlabeled-COCO images, which are out-of-domain data and have no inherent matches between captions and images in this setup.
	\vspace{-4mm}
	}
	\label{table:ablation_semi}	
\end{table}

%% file: 5_sec_appendix.tex

\subsection{Implementation Details}
We implement our neural networks by PyTorch~\cite{paszke2017automatic}.
We use the ResNet101 model~\cite{he2016deep} as a backbone CNN encoder for image input and initialize it with the weight pre-trained on ImageNet~\cite{russakovsky2015imagenet}.
For attention computation, we use spatial visual features of size $2048 \times 7 \times 7$ obtained from the last residual block of the backbone unless mentioned, \eg~NIC~\cite{vinyals2015show} and Att2in~\cite{rennie2017self}. 
For the other models, 
we use Faster R-CNN~\cite{ren2015faster} pre-trained on the Visual Genome~\cite{krishna2017visual} object dataset to extract 10 to 100 region features.
We set the {channel} size to be $1024$ for the hidden layers of LSTMs, $512$ for the attention layer, and $1024$ for the word embeddings.
For the inference stage, we empirically choose the beam size to be 3 when generating a caption, which shows the best performance.
We use a {mini}-batch of size 100 and the Adam optimizer~\cite{ba2015adam} for training with the hyper-parameters $lr {=} 5e^{-4}$, $b_1{=}0.9$, $b_2{=}0.999$.
We set $\lambda_x$ and $\lambda_y$ to be equal to $0.1$.

In order to avoid high complexity of the pseudo-labeling process, we do not search the pseudo-labels in the whole unpaired data, but in 
a 
$1\%$ of the unpaired data (yielding $1,000$ samples per a batch). 
Thus, the complexity of the pseudo-label retrieval for each mini-batch becomes $0.01\cdot O(BN),$ where $B$ denotes the mini-batch size, and $N$ the size of the unpaired dataset.
Since we also apply the label-noise handling, a plausible pseudo-label assignment is sufficient in helping the learning process even though the pseudo-label is not perfectly matching with the unpaired sample. 

\subsection{Connection with CycleGAN}
As a representative fully unpaired method, 
CycleGAN~\cite{zhu2017unpaired} would be a strong baseline,
which has been popularly used for unpaired distribution matching.
Since it is designed for the image-to-image translation, we describe the modifications to fit our task, so as to understand relative performance of our method over the CycleGAN baseline.
When applying the cycle-consistency on matching between images and captions, 
since the input modalities are totally different, we modify it to a translation problem over feature spaces as follows:
\begin{equation}
\centering\small
\begin{aligned}
    &\min_{F, G, \{T\}} \max_{D_{\{x,y\}}}  L_{cycle}(\{T\})\\ 
    &+\mathop{\mathbb{E}}\limits_{\mathbf{x}\sim p(\mathbf{x})}[\log(D_x(F(\mathbf{x}))) + \log(1-D_x(T_{v\rightarrow c}(F(\mathbf{x}))))]\\
    &+\mathop{\mathbb{E}}\limits_{y\sim p(y)}[\log(D_y(G(y))) + \log(1-D_y(T_{c\rightarrow v}(G(y))))],
\end{aligned}
\label{eqn:CycleGAN}
\end{equation}
where 
$\{T\}$ and $D_{\{x,y\}}$ denotes the feature translators and the discriminators for respective image and caption domains, and
\begin{equation}
\centering\small
\begin{aligned}
	L_{cycle}(\{T\}) = \mathbb{E}_{\mathbf{x}\sim p(\mathbf{x})}[\lVert T_{c\rightarrow v}(T_{v\rightarrow c}(F(\mathbf{x})))-F(\mathbf{x})\rVert_2]  
    \\+ \mathbb{E}_{y\sim p(y)}[\lVert T_{v\rightarrow c}(T_{c\rightarrow v}(G(y)))-G(y)\rVert_2].
\end{aligned}
\label{eqn:Cycle-Consistency}
\end{equation}
The discriminator $D$ is learned to distinguish whether the latent features are from the image or the caption distribution. 
This is different with our method, in that we distinguish the correct \emph{semantic match of pairs}.
We attempted training CycleGAN purely on unpaired datasets, but we were not able to train it; hence, we add the supervised loss
with the paired data.
This can be regarded 
as an improved CycleGAN and a fair setting with ours.

\section{Proof of Lemma 1}
Given any fixed generators $F$, $G$, and $\{T\}$, 
we are interested in maximizing the utility function $U$ in 
Eq.~(4)
over $D$.
The function $U$ can be rewritten as:
\begin{equation}
\begin{aligned}
U = &\E_{(\bz^x,\bz^y) \sim p(\bz^x,\bz^y)}\left[\log(D(\bz^x,\bz^y))\right]\\
+& \tfrac{1}{2}\E_{\bz^x \sim p(\bz^x)}\left[\log(1-D(\bz^x,T_{v \rightarrow c}(\bz^x)))\right]\\
+& \tfrac{1}{2}\E_{\bz^y \sim p(\bz^y)}\left[\log(1-D(T_{c \rightarrow v}(\bz^y),\bz^y)\right]\\
=& \E_{(\bz^x,\bz^y) \sim p(\bz^x,\bz^y)}\left[\log(D(\bz^x,\bz^y))\right]\\
+& \tfrac{1}{2}{\E}_{{\bz^x \sim p(\bz^x), \tilde\bz^y \sim \mathop{p}\limits_{v \rightarrow c}(\bz^y|\bz^x)}}\left[\log(1-D(\bz^x,\tilde\bz^y))\right]\\
+& \tfrac{1}{2}{\E}_{{\tilde\bz^x \sim \mathop{p}\limits_{c \rightarrow v}(\bz^x|\bz^y), \bz^y \sim p(\bz^y)}}\left[\log(1-D(\tilde\bz^x,\bz^y)\right]\\
=& \E_{(\bz^x,\bz^y) \sim p(\bz^x,\bz^y)}\left[\log(D(\bz^x,\bz^y))\right]\\
+& \tfrac{1}{2}{\E}_{{(\bz^x, \bz^y) \sim \mathop{p}\limits_{v \rightarrow c}(\bz^x, \bz^y)}}\left[\log(1-D(\bz^x,\bz^y))\right]\\
+& \tfrac{1}{2}{\E}_{{(\bz^x, \bz^y ) \sim \mathop{p}\limits_{c \rightarrow v}(\bz^x,\bz^y)}}\left[\log(1-D(\bz^x,\bz^y)\right]\\
=& \textstyle\int\int p(\bz^x,\bz^y)\log(D(\bz^x,\bz^y)) d\bz^x d\bz^y \\
+& \tfrac{1}{2} \textstyle\int p_{v \rightarrow c} (\bz^x,\bz^y) \log(1-D(\bz^x,\bz^y)) d\bz^x d\bz^y\\
+& \tfrac{1}{2} \textstyle\int p_{c \rightarrow v} (\bz^x, \bz^y)\log(1-D(\bz^x,\bz^y)) d\bz^x d\bz^y \\ 
=& \textstyle\int\int p(\bz^x,\bz^y)\log(D(\bz^x,\bz^y))  +\\
& p_{1/2}(\bz^x,\bz^y)\log(1-D(\bz^x,\bz^y)) d\bz^x d\bz^y.
\end{aligned}
\label{eq:U_alogblog}
\end{equation}
For any $(\alpha,\beta)\in \{\R\setminus\{0\}\}^2$ and $y\in[0,1]$, $f(y)=\alpha \log(y) + \beta \log(1{-}y)$ has a maximum at $y=\tfrac{\alpha}{\alpha+\beta}$, where \Eref{eq:U_alogblog} has the same form with $f(y)$.
This leads to $D^*(\bz^x,\bz^y)= \tfrac{p(\bz^x,\bz^y)}{p(\bz^x,\bz^y) + p_{1/2}(\bz^x,\bz^y)}$.
\hfill$\square$

\section{Proof of Lemma 2}
From the fact that 
\begin{equation}
\begin{aligned}
U =& \int\int p(\mathbf{z}^x,\mathbf{z}^y)\log(D(\mathbf{z}^x,\mathbf{z}^y)) d\mathbf{z}^x d\mathbf{z}^y \\
+& \int\int p_{1/2}(\mathbf{z}^x,\mathbf{z}^y)\log(1-D(\mathbf{z}^x,\mathbf{z}^y)) d\mathbf{z}^x d\mathbf{z}^y,
\end{aligned}
\end{equation}
derived in the proof of 
Lemma~1,
we can reformulate $V$ by setting $D = D^*(\mathbf{z}^x,\mathbf{z}^y)$ to maximize $U$ as:
\begin{equation}
\begin{aligned}
V = &\int\int p(\mathbf{z}^x,\mathbf{z}^y)\log(\frac{p(\mathbf{z}^x,\mathbf{z}^y)}{p(\mathbf{z}^x,\mathbf{z}^y) + p_{1/2}(\mathbf{z}^x,\mathbf{z}^y)}) d\mathbf{z}^x d\mathbf{z}^y \\
+ &\int\int p_{1/2}(\mathbf{z}^x,\mathbf{z}^y)\log(\frac{p_{1/2}(\mathbf{z}^x,\mathbf{z}^y)}{p(\mathbf{z}^x,\mathbf{z}^y) + p_{1/2}(\mathbf{z}^x,\mathbf{z}^y)}) d\mathbf{z}^x d\mathbf{z}^y
\end{aligned}
\label{eq:V_exp}
\end{equation}
\begin{equation}
\begin{aligned}
\phantom{V}=& \mathrm{KL}\left[p||\left(p+p_{1/2}\right)\right] + \mathrm{KL}\left[p_{1/2}||\left(p+p_{1/2}\right)\right]\\
=& \mathrm{KL}\left[p||\left(\tfrac{p+p_{1/2}}{2}\right)\right] + \mathrm{KL}\left[p_{1/2}||\left(\tfrac{p+p_{1/2}}{2}\right)\right] - \log4\qquad \\
=& 2\cdot \mathrm{JSD}(p||p_{1/2})  - \log4,\nonumber
\end{aligned}
\end{equation}
where $\mathrm{KL}$ is the Kullback-Leibler divergence, $\mathrm{JSD}$ is the Jensen-Shannon divergence between the two distributions, which is always non-negative and its unique optimum is achieved if and only if $p(\mathbf{z}^x,\mathbf{z}^y)=p_{1/2}(\mathbf{z}^x,\mathbf{z}^y)= \tfrac{p_{v\rightarrow c}(\mathbf{z}^x,\mathbf{z}^y)+p_{c\rightarrow v}(\mathbf{z}^x,\mathbf{z}^y)}{2}$.
Thereby, we have shown that $V{\geq} - \log4$ if and only if $p{=}p_{1/2}$.
Furthermore, for $p=p_{1/2}$, $D^*(\mathbf{z}^x,\mathbf{z}^y) = \tfrac{1}{2}$ according to 
Lemma~1.
By inspecting \Eref{eq:V_exp}, we see that
\begin{equation}
\begin{aligned}
V = & \E_p \left[\log(D^*)\right] + \E_{p_{1/2}} \left[\log(1-D^*)\right] \\
= & \E_p \left[\log\tfrac{1}{2}\right] + \E_{p_{1/2}} \left[\log\tfrac{1}{2}\right] = -\log{4}.
\end{aligned}
\end{equation}
This is the best possible value $V^* = - \log4$, reached only for $p=p_{1/2}$.

Next, we show that, at the optimal point, $V^*$ guarantees at least the marginal distribution consistency, \ie~$p(\mathbf{z}^y)=p_{c\rightarrow v}(\mathbf{z}^y)=p_{v\rightarrow c}(\mathbf{z}^y)$ and $p(\mathbf{z}^x)=p_{c\rightarrow v}(\mathbf{z}^x)=p_{v\rightarrow c}(\mathbf{z}^x)$.
Given $p(\mathbf{z}^x,\mathbf{z}^y)=p_{1/2}(\mathbf{z}^x,\mathbf{z}^y)$, we first take the marginalization with respect to $\mathbf{z}^x$ on both sides of $p(\mathbf{z}^x,\mathbf{z}^y)=p_{1/2}(\mathbf{z}^x,\mathbf{z}^y)$.
Since we have $p_{v\rightarrow c}(\mathbf{z}^x,\mathbf{z}^y) = p(\mathbf{z}^x)\cdot p_{v\rightarrow c}(\mathbf{z}^y|\mathbf{z}^x)$ and $p_{c\rightarrow v}(\mathbf{z}^x,\mathbf{z}^y) = p(\mathbf{z}^y)\cdot p_{c\rightarrow v}(\mathbf{z}^x|\mathbf{z}^y)$, the marginalization results in 
\begin{equation}
\begin{aligned}
\int p(\mathbf{z}^x,\mathbf{z}^y) d\bz^x = \tfrac{1}{2}\int p_{v\rightarrow c}(\mathbf{z}^x,\mathbf{z}^y) + p_{c\rightarrow v}(\mathbf{z}^x,\mathbf{z}^y) d\bz^x,
\end{aligned}
\end{equation}
which is equal to $p(\mathbf{z}^y) = \tfrac{1}{2}p_{v\rightarrow c}(\mathbf{z}^y) + \tfrac{1}{2}p(\mathbf{z}^y)$, \ie~$p(\mathbf{z}^y) = p_{v\rightarrow c}(\mathbf{z}^y) = p_{c\rightarrow v}(\mathbf{z}^y)$.
We can show $p(\mathbf{z}^x) = p_{v\rightarrow c}(\mathbf{z}^x) = p_{c\rightarrow v}(\mathbf{z}^x)$ analogously.
\hfill$\square$

\section{Proof of Theorem 1}
In 
Eq.~(12),
$\mathrm{KL}\left[p(\mathbf{z}^x|\mathbf{z}^y)||p_{c\rightarrow v}(\mathbf{z}^x|\mathbf{z}^y)\right]$ 
is always non-negative and zero if and only if $p(\mathbf{z}^x|\mathbf{z}^y)= p_{c\rightarrow v}(\mathbf{z}^x|\mathbf{z}^y)$.
The same can be applied to show $p(\mathbf{z}^y|\mathbf{z}^x)= p_{v\rightarrow c}(\mathbf{z}^y|\mathbf{z}^x)$ with $\mathrm{KL}\left[p(\mathbf{z}^y|\mathbf{z}^x)||p_{v\rightarrow c}(\mathbf{z}^y|\mathbf{z}^x)\right]$, which we called the conditional distribution match.

Furthermore, from 
Lemmas~1 and 2,
we have 
$p(\mathbf{z}^x,\mathbf{z}^y)=p_{1/2}(\mathbf{z}^x,\mathbf{z}^y)$ and the marginal distribution consistency $p(\mathbf{z}^y)=p_{c\rightarrow v}(\mathbf{z}^y)=p_{v\rightarrow c}(\mathbf{z}^y)$ and $p(\mathbf{z}^x)=p_{c\rightarrow v}(\mathbf{z}^x)=p_{v\rightarrow c}(\mathbf{z}^x)$ at a  global equilibrium point of ${U}(\theta, \{T\},D)$. Then, the rest follows the same logic with \cite{chongxuan2017triple}.
With the marginal and conditional distribution consistencies, it is easy to verify that $p(\mathbf{z}^x,\mathbf{z}^y) = p_{c\rightarrow v}(\mathbf{z}^x,\mathbf{z}^y)$ and $p(\mathbf{z}^x,\mathbf{z}^y)  = p_{v\rightarrow c}(\mathbf{z}^x,\mathbf{z}^y)$. Then, $p(\mathbf{z}^x,\mathbf{z}^y)=p_{1/2}(\mathbf{z}^x,\mathbf{z}^y)$ implies $p(\mathbf{z}^x,\mathbf{z}^y) = p_{c\rightarrow v}(\mathbf{z}^x,\mathbf{z}^y)= p_{v\rightarrow c}(\mathbf{z}^x,\mathbf{z}^y)$, which
is a global equilibrium of 
Eq.~(12).
\hfill$\square$

%% file: TIP.bbl
\begin{thebibliography}{10}\itemsep=-1pt

\bibitem{anderson2016spice}
P.~Anderson, B.~Fernando, M.~Johnson, and S.~Gould.
\newblock Spice: Semantic propositional image caption evaluation.
\newblock In {\em European Conference on Computer Vision (ECCV)}. Springer,
  2016.

\bibitem{anderson2017bottom}
P.~Anderson, X.~He, C.~Buehler, D.~Teney, M.~Johnson, S.~Gould, and L.~Zhang.
\newblock Bottom-up and top-down attention for image captioning and vqa.
\newblock In {\em {IEEE} Conference on Computer Vision and Pattern Recognition
  (CVPR)}, 2018.

\bibitem{anne2016deep}
L.~Anne~Hendricks, S.~Venugopalan, M.~Rohrbach, R.~Mooney, K.~Saenko, and
  T.~Darrell.
\newblock Deep compositional captioning: Describing novel object categories
  without paired training data.
\newblock In {\em {IEEE} Conference on Computer Vision and Pattern Recognition
  (CVPR)}, 2016.

\bibitem{artetxe2018unsupervised}
M.~Artetxe, G.~Labaka, E.~Agirre, and K.~Cho.
\newblock Unsupervised neural machine translation.
\newblock In {\em International Conference on Learning Representations (ICLR)},
  2018.

\bibitem{ben2021unpaired}
H.~Ben, Y.~Pan, Y.~Li, T.~Yao, R.~Hong, M.~Wang, and T.~Mei.
\newblock Unpaired image captioning with semantic-constrained self-learning.
\newblock {\em IEEE Transactions on Multimedia}, 2021.

\bibitem{berthelot2020remixmatch}
D.~Berthelot, N.~Carlini, E.~D. Cubuk, A.~Kurakin, K.~Sohn, H.~Zhang, and
  C.~Raffel.
\newblock Remixmatch: Semi-supervised learning with distribution matching and
  augmentation anchoring.
\newblock In {\em International Conference on Learning Representations (ICLR)},
  2020.

\bibitem{berthelot2019mixmatch}
D.~Berthelot, N.~Carlini, I.~Goodfellow, N.~Papernot, A.~Oliver, and C.~A.
  Raffel.
\newblock Mixmatch: A holistic approach to semi-supervised learning.
\newblock In {\em Advances in Neural Information Processing Systems (NIPS)},
  2019.

\bibitem{chapelle2009semi}
O.~Chapelle, B.~Scholkopf, and A.~Zien.
\newblock Semi-supervised learning.
\newblock {\em {IEEE} Transactions on Neural Networks}, 20(3):542--542, 2009.

\bibitem{chen2017show}
T.-H. Chen, Y.-H. Liao, C.-Y. Chuang, W.-T. Hsu, J.~Fu, and M.~Sun.
\newblock Show, adapt and tell: Adversarial training of cross-domain image
  captioner.
\newblock In {\em {IEEE} International Conference on Computer Vision (ICCV)},
  2017.

\bibitem{chen2021self}
X.~Chen, M.~Jiang, and Q.~Zhao.
\newblock Self-distillation for few-shot image captioning.
\newblock In {\em IEEE Winter Conference on Applications of Computer Vision
  (WACV)}, 2021.

\bibitem{cho2022mcdal}
J.~W. Cho, D.-J. Kim, Y.~Jung, and I.~S. Kweon.
\newblock Mcdal: Maximum classifier discrepancy for active learning.
\newblock {\em IEEE transactions on neural networks and learning systems},
  2022.

\bibitem{cho2014learning}
K.~Cho, B.~Van~Merri{\"e}nboer, C.~Gulcehre, D.~Bahdanau, F.~Bougares,
  H.~Schwenk, and Y.~Bengio.
\newblock Learning phrase representations using rnn encoder-decoder for
  statistical machine translation.
\newblock In {\em Conference on Empirical Methods in Natural Language
  Processing (EMNLP)}, 2014.

\bibitem{choi2018contextually}
J.~Choi, T.-H. Oh, and I.~S. Kweon.
\newblock Contextually customized video summaries via natural language.
\newblock In {\em IEEE Winter Conference on Applications of Computer Vision
  (WACV)}, 2018.

\bibitem{chongxuan2017triple}
L.~Chongxuan, T.~Xu, J.~Zhu, and B.~Zhang.
\newblock Triple generative adversarial nets.
\newblock In {\em Advances in Neural Information Processing Systems (NIPS)},
  2017.

\bibitem{cornia2020meshed}
M.~Cornia, M.~Stefanini, L.~Baraldi, and R.~Cucchiara.
\newblock Meshed-memory transformer for image captioning.
\newblock In {\em {IEEE} Conference on Computer Vision and Pattern Recognition
  (CVPR)}, 2020.

\bibitem{denkowski2014meteor}
M.~Denkowski and A.~Lavie.
\newblock Meteor universal: Language specific translation evaluation for any
  target language.
\newblock In {\em The workshop on statistical machine translation}, 2014.

\bibitem{feng2018unsupervised}
Y.~Feng, L.~Ma, W.~Liu, and J.~Luo.
\newblock Unsupervised image captioning.
\newblock In {\em {IEEE} Conference on Computer Vision and Pattern Recognition
  (CVPR)}, 2019.

\bibitem{gan2017triangle}
Z.~Gan, L.~Chen, W.~Wang, Y.~Pu, Y.~Zhang, H.~Liu, C.~Li, and L.~Carin.
\newblock Triangle generative adversarial networks.
\newblock In {\em Advances in Neural Information Processing Systems (NIPS)},
  2017.

\bibitem{goodfellow2014generative}
I.~Goodfellow, J.~Pouget-Abadie, M.~Mirza, B.~Xu, D.~Warde-Farley, S.~Ozair,
  A.~Courville, and Y.~Bengio.
\newblock Generative adversarial nets.
\newblock In {\em Advances in Neural Information Processing Systems (NIPS)},
  2014.

\bibitem{gu2018unpaired}
J.~Gu, S.~Joty, J.~Cai, and G.~Wang.
\newblock Unpaired image captioning by language pivoting.
\newblock In {\em European Conference on Computer Vision (ECCV)}, 2018.

\bibitem{gu2019unpaired}
J.~Gu, S.~Joty, J.~Cai, H.~Zhao, X.~Yang, and G.~Wang.
\newblock Unpaired image captioning via scene graph alignments.
\newblock In {\em {IEEE} International Conference on Computer Vision (ICCV)},
  2019.

\bibitem{guo2020recurrent}
D.~Guo, Y.~Wang, P.~Song, and M.~Wang.
\newblock Recurrent relational memory network for unsupervised image
  captioning.
\newblock In {\em International Joint Conference on Artificial Intelligence
  (IJCAI)}, 2020.

\bibitem{hinton2015distilling}
G.~Hinton, O.~Vinyals, and J.~Dean.
\newblock Distilling the knowledge in a neural network.
\newblock {\em arXiv preprint arXiv:1503.02531}, 2015.

\bibitem{hu2022scaling}
X.~Hu, Z.~Gan, J.~Wang, Z.~Yang, Z.~Liu, Y.~Lu, and L.~Wang.
\newblock Scaling up vision-language pre-training for image captioning.
\newblock In {\em {IEEE} Conference on Computer Vision and Pattern Recognition
  (CVPR)}, 2022.

\bibitem{huang2019attention}
L.~Huang, W.~Wang, J.~Chen, and X.-Y. Wei.
\newblock Attention on attention for image captioning.
\newblock In {\em {IEEE} International Conference on Computer Vision (ICCV)},
  2019.

\bibitem{johnson2016densecap}
J.~Johnson, A.~Karpathy, and L.~Fei-Fei.
\newblock Densecap: Fully convolutional localization networks for dense
  captioning.
\newblock In {\em {IEEE} Conference on Computer Vision and Pattern Recognition
  (CVPR)}, 2016.

\bibitem{karpathy2015deep}
A.~Karpathy and L.~Fei-Fei.
\newblock Deep visual-semantic alignments for generating image descriptions.
\newblock In {\em {IEEE} Conference on Computer Vision and Pattern Recognition
  (CVPR)}, 2015.

\bibitem{kim2021single}
D.-J. Kim, J.~W. Cho, J.~Choi, Y.~Jung, and I.~S. Kweon.
\newblock Single-modal entropy based active learning for visual question
  answering.
\newblock In {\em British Machine Vision Conference (BMVC)}, 2021.

\bibitem{kim2019dense}
D.-J. Kim, J.~Choi, T.-H. Oh, and I.~S. Kweon.
\newblock Dense relational captioning: Triple-stream networks for
  relationship-based captioning.
\newblock In {\em Proceedings of the IEEE Conference on Computer Vision and
  Pattern Recognition}, pages 6271--6280, 2019.

\bibitem{kim2019image}
D.-J. Kim, J.~Choi, T.-H. Oh, and I.~S. Kweon.
\newblock Image captioning with very scarce supervised data: Adversarial
  semi-supervised learning approach.
\newblock In {\em Proceedings of the 2019 Conference on Empirical Methods in
  Natural Language Processing and the 9th International Joint Conference on
  Natural Language Processing (EMNLP-IJCNLP)}, 2019.

\bibitem{kim2018disjoint}
D.-J. Kim, J.~Choi, T.-H. Oh, Y.~Yoon, and I.~S. Kweon.
\newblock Disjoint multi-task learning between heterogeneous human-centric
  tasks.
\newblock In {\em 2018 IEEE Winter Conference on Applications of Computer
  Vision (WACV)}, pages 1699--1708. IEEE, 2018.

\bibitem{kim2021dense}
D.-J. Kim, T.-H. Oh, J.~Choi, and I.~S. Kweon.
\newblock Dense relational image captioning via multi-task triple-stream
  networks.
\newblock {\em IEEE Transactions on Pattern Analysis and Machine Intelligence},
  2021.

\bibitem{kim2018textual}
J.~Kim, A.~Rohrbach, T.~Darrell, J.~Canny, and Z.~Akata.
\newblock Textual explanations for self-driving vehicles.
\newblock In {\em European Conference on Computer Vision (ECCV)}, 2018.

\bibitem{kim2017learning}
T.~Kim, M.~Cha, H.~Kim, J.~K. Lee, and J.~Kim.
\newblock Learning to discover cross-domain relations with generative
  adversarial networks.
\newblock In {\em International Conference on Machine Learning (ICML)}, 2017.

\bibitem{krasin2017openimages}
I.~Krasin, T.~Duerig, N.~Alldrin, V.~Ferrari, S.~Abu-El-Haija, A.~Kuznetsova,
  H.~Rom, J.~Uijlings, S.~Popov, A.~Veit, et~al.
\newblock Openimages: A public dataset for large-scale multi-label and
  multi-class image classification.
\newblock {\em Dataset available from https://github. com/openimages}, 2:3,
  2017.

\bibitem{krishna2017visual}
R.~Krishna, Y.~Zhu, O.~Groth, J.~Johnson, K.~Hata, J.~Kravitz, S.~Chen,
  Y.~Kalantidis, L.-J. Li, D.~A. Shamma, et~al.
\newblock Visual genome: Connecting language and vision using crowdsourced
  dense image annotations.
\newblock {\em International Journal of Computer Vision (IJCV)}, 123(1):32--73,
  2017.

\bibitem{krizhevsky2012imagenet}
A.~Krizhevsky, I.~Sutskever, and G.~E. Hinton.
\newblock Imagenet classification with deep convolutional neural networks.
\newblock In {\em Advances in Neural Information Processing Systems (NIPS)},
  2012.

\bibitem{kuo2020featmatch}
C.~Kuo, C.~Ma, J.~Huang, and Z.~Kira.
\newblock Featmatch: Feature-based augmentation for semi-supervised learning.
\newblock In {\em European Conference on Computer Vision (ECCV)}, 2020.

\bibitem{kuo2022beyond}
C.-W. Kuo and Z.~Kira.
\newblock Beyond a pre-trained object detector: Cross-modal textual and visual
  context for image captioning.
\newblock In {\em {IEEE} Conference on Computer Vision and Pattern Recognition
  (CVPR)}, 2022.

\bibitem{laina2019towards}
I.~Laina, C.~Rupprecht, and N.~Navab.
\newblock Towards unsupervised image captioning with shared multimodal
  embeddings.
\newblock In {\em {IEEE} International Conference on Computer Vision (ICCV)},
  2019.

\bibitem{lee2013pseudo}
D.-H. Lee.
\newblock Pseudo-label: The simple and efficient semi-supervised learning
  method for deep neural networks.
\newblock In {\em Workshop on Challenges in Representation Learning, ICML},
  2013.

\bibitem{lee2017cleannet}
K.-H. Lee, X.~He, L.~Zhang, and L.~Yang.
\newblock Cleannet: Transfer learning for scalable image classifier training
  with label noise.
\newblock In {\em {IEEE} Conference on Computer Vision and Pattern Recognition
  (CVPR)}, 2018.

\bibitem{lin2004rouge}
C.-Y. Lin.
\newblock Rouge: A package for automatic evaluation of summaries.
\newblock {\em Text Summarization Branches Out}, 2004.

\bibitem{lin2014microsoft}
T.-Y. Lin, M.~Maire, S.~Belongie, J.~Hays, P.~Perona, D.~Ramanan,
  P.~Doll{\'a}r, and C.~L. Zitnick.
\newblock Microsoft coco: Common objects in context.
\newblock In {\em European Conference on Computer Vision (ECCV)}. Springer,
  2014.

\bibitem{liu2017unsupervised}
M.-Y. Liu, T.~Breuel, and J.~Kautz.
\newblock Unsupervised image-to-image translation networks.
\newblock In {\em Advances in Neural Information Processing Systems (NIPS)},
  2017.

\bibitem{liu2018show}
X.~Liu, H.~Li, J.~Shao, D.~Chen, and X.~Wang.
\newblock Show, tell and discriminate: Image captioning by self-retrieval with
  partially labeled data.
\newblock In {\em European Conference on Computer Vision (ECCV)}, 2018.

\bibitem{lu2016visual}
C.~Lu, R.~Krishna, M.~Bernstein, and L.~Fei-Fei.
\newblock Visual relationship detection with language priors.
\newblock In {\em European Conference on Computer Vision (ECCV)}. Springer,
  2016.

\bibitem{lu2018neural}
J.~Lu, J.~Yang, D.~Batra, and D.~Parikh.
\newblock Neural baby talk.
\newblock In {\em {IEEE} Conference on Computer Vision and Pattern Recognition
  (CVPR)}, 2018.

\bibitem{meng2022object}
Z.~Meng, D.~Yang, X.~Cao, A.~Shah, and S.-N. Lim.
\newblock Object-centric unsupervised image captioning.
\newblock In {\em European Conference on Computer Vision (ECCV)}, 2022.

\bibitem{miyato2018virtual}
T.~Miyato, S.-i. Maeda, M.~Koyama, and S.~Ishii.
\newblock Virtual adversarial training: a regularization method for supervised
  and semi-supervised learning.
\newblock {\em {IEEE} Transactions on Pattern Analysis and Machine Intelligence
  (TPAMI)}, 41(8):1979--1993, 2018.

\bibitem{nguyen2022grit}
V.-Q. Nguyen, M.~Suganuma, and T.~Okatani.
\newblock Grit: Faster and better image captioning transformer using dual
  visual features.
\newblock In {\em European Conference on Computer Vision (ECCV)}, 2022.

\bibitem{oh2022daso}
Y.~Oh, D.-J. Kim, and I.~S. Kweon.
\newblock Daso: Distribution-aware semantics-oriented pseudo-label for
  imbalanced semi-supervised learning.
\newblock In {\em {IEEE} Conference on Computer Vision and Pattern Recognition
  (CVPR)}, 2022.

\bibitem{pan2020x}
Y.~Pan, T.~Yao, Y.~Li, and T.~Mei.
\newblock X-linear attention networks for image captioning.
\newblock In {\em {IEEE} Conference on Computer Vision and Pattern Recognition
  (CVPR)}, 2020.

\bibitem{papineni2002bleu}
K.~Papineni, S.~Roukos, T.~Ward, and W.-J. Zhu.
\newblock Bleu: a method for automatic evaluation of machine translation.
\newblock In {\em Association for Computational Linguistics (ACL)}. Association
  for Computational Linguistics, 2002.

\bibitem{pennington2014glove}
J.~Pennington, R.~Socher, and C.~D. Manning.
\newblock Glove: Global vectors for word representation.
\newblock In {\em Proceedings of the 2014 conference on empirical methods in
  natural language processing (EMNLP)}, 2014.

\bibitem{ren2015faster}
S.~Ren, K.~He, R.~Girshick, and J.~Sun.
\newblock Faster {R-CNN}: Towards real-time object detection with region
  proposal networks.
\newblock In {\em Advances in Neural Information Processing Systems (NIPS)},
  2015.

\bibitem{rennie2017self}
S.~J. Rennie, E.~Marcheret, Y.~Mroueh, J.~Ross, and V.~Goel.
\newblock Self-critical sequence training for image captioning.
\newblock In {\em {IEEE} Conference on Computer Vision and Pattern Recognition
  (CVPR)}, 2017.

\bibitem{shalev2014understanding}
S.~Shalev-Shwartz and S.~Ben-David.
\newblock {\em Understanding machine learning: From theory to algorithms}.
\newblock Cambridge university press, 2014.

\bibitem{shi2018transductive}
W.~Shi, Y.~Gong, C.~Ding, Z.~MaXiaoyu~Tao, and N.~Zheng.
\newblock Transductive semi-supervised deep learning using min-max features.
\newblock In {\em European Conference on Computer Vision (ECCV)}, 2018.

\bibitem{shin2021labor}
I.~Shin, D.-J. Kim, J.~W. Cho, S.~Woo, K.~Park, and I.~S. Kweon.
\newblock Labor: Labeling only if required for domain adaptive semantic
  segmentation.
\newblock In {\em {IEEE} International Conference on Computer Vision (ICCV)},
  2021.

\bibitem{sohn2020fixmatch}
K.~Sohn, D.~Berthelot, C.~Li, Z.~Zhang, N.~Carlini, E.~D. Cubuk, A.~Kurakin,
  H.~Zhang, and C.~Raffel.
\newblock Fixmatch: Simplifying semi-supervised learning with consistency and
  confidence.
\newblock In {\em Advances in Neural Information Processing Systems (NIPS)},
  2020.

\bibitem{tarvainen2017mean}
A.~Tarvainen and H.~Valpola.
\newblock Mean teachers are better role models: Weight-averaged consistency
  targets improve semi-supervised deep learning results.
\newblock In {\em Advances in Neural Information Processing Systems (NIPS)},
  2017.

\bibitem{thomee2016yfcc100m}
B.~Thomee, D.~A. Shamma, G.~Friedland, B.~Elizalde, K.~Ni, D.~Poland, D.~Borth,
  and L.-J. Li.
\newblock Yfcc100m: the new data in multimedia research.
\newblock {\em Communications of the ACM}, 59(2):64--73, 2016.

\bibitem{utiyama2007comparison}
M.~Utiyama and H.~Isahara.
\newblock A comparison of pivot methods for phrase-based statistical machine
  translation.
\newblock In {\em Conference of the North American Chapter of the Association
  for Computational Linguistics: Human Language Technologies (NAACL-HLT)},
  2007.

\bibitem{vedantam2015cider}
R.~Vedantam, C.~Lawrence~Zitnick, and D.~Parikh.
\newblock Cider: Consensus-based image description evaluation.
\newblock In {\em {IEEE} Conference on Computer Vision and Pattern Recognition
  (CVPR)}, 2015.

\bibitem{venugopalan2017captioning}
S.~Venugopalan, L.~Anne~Hendricks, M.~Rohrbach, R.~Mooney, T.~Darrell, and
  K.~Saenko.
\newblock Captioning images with diverse objects.
\newblock In {\em {IEEE} Conference on Computer Vision and Pattern Recognition
  (CVPR)}, 2017.

\bibitem{vinyals2015show}
O.~Vinyals, A.~Toshev, S.~Bengio, and D.~Erhan.
\newblock Show and tell: A neural image caption generator.
\newblock In {\em {IEEE} Conference on Computer Vision and Pattern Recognition
  (CVPR)}, 2015.

\bibitem{wang2018look}
X.~Wang, W.~Xiong, H.~Wang, and W.~Yang~Wang.
\newblock Look before you leap: Bridging model-free and model-based
  reinforcement learning for planned-ahead vision-and-language navigation.
\newblock In {\em European Conference on Computer Vision (ECCV)}, 2018.

\bibitem{wang2018iterative}
Y.~Wang, W.~Liu, X.~Ma, J.~Bailey, H.~Zha, L.~Song, and S.-T. Xia.
\newblock Iterative learning with open-set noisy labels.
\newblock In {\em {IEEE} Conference on Computer Vision and Pattern Recognition
  (CVPR)}, 2018.

\bibitem{wu2017ai}
J.~Wu, H.~Zheng, B.~Zhao, Y.~Li, B.~Yan, R.~Liang, W.~Wang, S.~Zhou, G.~Lin,
  Y.~Fu, et~al.
\newblock Ai challenger: A large-scale dataset for going deeper in image
  understanding.
\newblock {\em arXiv preprint arXiv:1711.06475}, 2017.

\bibitem{xu2015show}
K.~Xu, J.~Ba, R.~Kiros, K.~Cho, A.~Courville, R.~Salakhudinov, R.~Zemel, and
  Y.~Bengio.
\newblock Show, attend and tell: Neural image caption generation with visual
  attention.
\newblock In {\em International Conference on Machine Learning (ICML)}, 2015.

\bibitem{Yang_2017_CVPR}
L.~Yang, K.~Tang, J.~Yang, and L.-J. Li.
\newblock Dense captioning with joint inference and visual context.
\newblock In {\em {IEEE} Conference on Computer Vision and Pattern Recognition
  (CVPR)}, 2017.

\bibitem{yang2018shuffle}
X.~Yang, H.~Zhang, and J.~Cai.
\newblock Shuffle-then-assemble: learning object-agnostic visual relationship
  features.
\newblock In {\em European Conference on Computer Vision (ECCV)}, 2018.

\bibitem{young2014image}
P.~Young, A.~Lai, M.~Hodosh, and J.~Hockenmaier.
\newblock From image descriptions to visual denotations: New similarity metrics
  for semantic inference over event descriptions.
\newblock {\em Association for Computational Linguistics (ACL)}, 2014.

\bibitem{zhang2015character}
X.~Zhang, J.~Zhao, and Y.~LeCun.
\newblock Character-level convolutional networks for text classification.
\newblock In {\em Advances in Neural Information Processing Systems (NIPS)},
  2015.

\bibitem{zhang2018joint}
Z.~Zhang, S.~Liu, M.~Li, M.~Zhou, and E.~Chen.
\newblock Joint training for neural machine translation models with monolingual
  data.
\newblock In {\em AAAI Conference on Artificial Intelligence (AAAI)}, 2018.

\bibitem{zhou2004learning}
D.~Zhou, O.~Bousquet, T.~N. Lal, J.~Weston, and B.~Sch{\"o}lkopf.
\newblock Learning with local and global consistency.
\newblock In {\em Advances in Neural Information Processing Systems (NIPS)},
  2004.

\bibitem{zhu2017unpaired}
J.-Y. Zhu, T.~Park, P.~Isola, and A.~A. Efros.
\newblock Unpaired image-to-image translation using cycle-consistent
  adversarial networks.
\newblock In {\em {IEEE} International Conference on Computer Vision (ICCV)},
  2017.

\bibitem{zhu2022unpaired}
P.~Zhu, X.~Wang, Y.~Luo, Z.~Sun, W.-S. Zheng, Y.~Wang, and C.~Chen.
\newblock Unpaired image captioning by image-level weakly-supervised visual
  concept recognition.
\newblock {\em arXiv preprint arXiv:2203.03195}, 2022.

\end{thebibliography}
